\documentclass[nojss]{jss}
\pdfoutput=1
\usepackage[usenames,dvipsnames]{xcolor}

\usepackage{thumbpdf,lmodern}
\usepackage{orcidlink}
\usepackage{amsmath,amssymb,bbold, xfrac}

\usepackage{multirow}
\usepackage{caption}
\usepackage{subcaption} % for subfigure
\usepackage{float}

\usepackage{tikz}
\usetikzlibrary{shapes.geometric, arrows, backgrounds, scopes}
\usepackage{pgfplots}
\pgfplotsset{width=6.75cm, compat=newest}
\usepackage[utf8]{inputenc}
\DeclareUnicodeCharacter{2212}{−}
\usepgfplotslibrary{groupplots,dateplot}
\usetikzlibrary{patterns,shapes.arrows}

\usepackage{algorithm}
\usepackage{algpseudocode} % algorithmicx package

\usepackage{framed}
\usepackage{marginnote}

\newcommand{\hess}{\mathbf{H}}

\newcommand{\vb}{\mathbf{v}}
\newcommand{\ub}{\mathbf{u}}
\newcommand{\yb}{\mathbf{y}}
\newcommand{\xb}{\mathbf{x}}

\newcommand{\jac}{\mathbf{J}}
\newcommand{\hessian}{\mathbf{H}}
\newcommand{\thetab}{\boldsymbol{\theta}}

\newcommand{\simulator}{g}
\newcommand{\region}{B_{d,\epsilon}}
\newcommand{\indicator}[1]{\mathbb{1}_{#1}}

\newcommand{\data}{\mathbf{y_0}}
\newcommand{\accregioni}{C^i_{\epsilon}}
\newcommand{\accregionitheta}{C^i_{\epsilon, \theta_i^*}}
\newcommand{\accregionihat}{\hat{C}^i_{\epsilon}}

\newcommand{\R}{\mathbb{R}}

\DeclareMathOperator*{\argmin}{argmin}

\author{Vasilis Gkolemis~\orcidlink{0000-0002-2636-0245}\\ATHENA RC \And
  Michael Gutmann~\orcidlink{0000-0002-5329-9910}\\University of Edinburgh \And
  Henri Pesonen~\orcidlink{0000-0003-4500-2926}\\University of Oslo}

\Plainauthor{Vasilis Gkolemis, Michael Gutmann, Henri Pesonen}

\title{An Extendable \proglang{Python} Implementation of Robust Optimisation Monte Carlo}
\Plaintitle{An Extendable Python Implementation of ROMC}
\Shorttitle{An Extendable ROMC implementation}

\Abstract{Performing inference in statistical models with an
  intractable likelihood is challenging, therefore, most
  likelihood-free inference (LFI) methods encounter accuracy and
  efficiency limitations. In this paper, we present the implementation
  of the LFI method Robust Optimisation Monte Carlo (ROMC) in the
  \proglang{Python} package \pkg{ELFI}. ROMC is a novel and efficient
  (highly-parallelizable) LFI framework that provides accurate
  weighted samples from the posterior. Our implementation can be used
  in two ways. First, a scientist may use it as an out-of-the-box LFI
  algorithm; we provide an easy-to-use API harmonized with the
  principles of \pkg{ELFI}, enabling effortless comparisons with the
  rest of the methods included in the package. Additionally, we have
  carefully split ROMC into isolated components for supporting
  extensibility. A researcher may experiment with novel method(s) for
  solving part(s) of ROMC without reimplementing everything from
  scratch. In both scenarios, the ROMC parts can run in a
  fully-parallelized manner, exploiting all CPU cores. We also provide
  helpful functionalities for (i) inspecting the inference process and
  (ii) evaluating the obtained samples. Finally, we test the
  robustness of our implementation on some typical LFI examples.}

\Plainkeywords{Bayesian inference, implicit models, likelihood-free, Python, ELFI}

\Address{
  Vasilis Gkolemis\\
  Information Management Systems Institute (IMSI)\\
  ATHENA Research and Innovation Center\\
  Athens, Greece\\
  E-mail: \email{vgkolemis@athenarc.gr}\\
  URL: \url{https://givasile.github.io}
}

\begin{document}

%% -- Introduction -------------------------------------------------------------
\section[Introduction]{Introduction}
\label{sec:intro}

Simulator-based models are particularly captivating due to the
modeling freedom they provide. In essence, any data generating
mechanism that can be written as a finite set of algorithmic steps can
be programmed as a simulator-based model. Hence, these models are
often used to model physical phenomena in the natural sciences such
as, e.g., genetics, epidemiology or neuroscience~\citet{gutmann2016,
  lintusaari2017, sisson2018, cranmer2020}. In simulator-based models,
it is feasible to generate samples using the simulator but is
infeasible to evaluate the likelihood function. The intractability of
the likelihood makes the so-called likelihood-free inference (LFI),
i.e., the approximation of the posterior distribution without using
the likelihood function, particularly challenging.

Optimization Monte Carlo (OMC), proposed by~\citet{Meeds2015}, is a
novel LFI approach. The central idea is to convert the stochastic
data-generating mechanism into a set of deterministic optimization
processes. Afterwards, \citet{Forneron2016} described a similar method
under the name `reverse sampler'. In their work,~\citet{Ikonomov2019}
identified some critical limitations of OMC, so they proposed Robust
OMC (ROMC) an improved version of OMC with appropriate modifications.

In this paper, we present the implementation of ROMC at the
\proglang{Python} package \pkg{ELFI} (\pkg{Engine for likelihood-free
  inference})~\citet{1708.00707}. The implementation has been designed
to facilitate extensibility. ROMC is an LFI framework; it defines a
sequence of algorithmic steps for approximating the posterior without
enforcing a specific algorithm for each step. Therefore, a researcher
may use ROMC as the backbone method and apply novel algorithms to each
separate step. For being a ready-to-use LFI
method,~\citet{Ikonomov2019} propose a particular (default) algorithm
for each step, but this choice is by no means restrictive. We have
designed our software for facilitating such experimentation.

To the best of our knowledge, this is the first implementation of the
ROMC inference method to a generic LFI framework. We organize the
illustration and the evaluation of our implementation in three
steps. First, for securing that our implementation is accurate, we
test it against an artificial example with a tractable likelihood. The
artificial example also serves as a step-by-step guide for showcasing
the various functionalities of our implementation. Second, we use the
second-order moving average (MA2) example~\citet{Marin2012} from the
\pkg{ELFI} package, using as ground truth the samples obtained with
Rejection ABC~\citet{lintusaari2017} using a very high number of
samples. Finally, we present the execution times of ROMC, measuring
the speed-up achieved by the parallel version of the implementation.

The code of the implementation is available at the official \pkg{ELFI}
\href{https://github.com/elfi-dev/elfi}{repository}. Apart from the
examples presented in the paper, there are five \pkg{Google
  Colab}~\citet{Bisong2019} notebooks available online, with
end-to-end examples illustrating how to: (i)
\href{https://colab.research.google.com/drive/1lGRp0XrNfZ64NN0ASB_tYEKowXwlveDC?usp=sharing}{use ROMC
  on a synthetic \(1D\) example}, (ii)
\href{https://colab.research.google.com/drive/1Fof_WmCi1YizzSI_63aEsbLXsno5gSZ3?usp=sharing}{use ROMC
  on a synthetic \(2D\) example}, (iii)
\href{https://colab.research.google.com/drive/1nkdACQ370SSc0KB1bHv4sBRaxMlMqoNH?usp=sharing}{use ROMC
  on the Moving Average example}, (iv)
\href{https://colab.research.google.com/drive/1_jHVxPSH3XcNOORZJpLU0SPzs0PF8CQ5?usp=sharing}{extend ROMC with a Neural Network as a surrogate model}, (v)
\href{https://colab.research.google.com/drive/1RzB-V1QueP1y1nyzv_VOqR1nVz3DUH3v?usp=sharing}{extend ROMC with a custom proposal region module}.

% \clearpage
\section[Background]{Background}

We first give a short introduction to simulator-based models, we then
focus on OMC and its robust version, ROMC, and we, finally, introduce
\pkg{ELFI}, the \proglang{Python} package used for the implementation.

\subsection{Simulator-based models and likelihood-free inference}

An implicit or simulator-based model is a parameterized stochastic
data generating mechanism, where we can sample data points but we
cannot evaluate the likelihood. Formally, a simulator-based model is a
parameterized family of probability density functions
\(\{ p(\yb|\thetab)\}_{\thetab}\) whose closed-form is either unknown
or computationally intractable. In these cases, we can only access the
simulator \( m_r(\thetab) \), i.e., the black-box mechanism (computer
code) that generates samples \(\yb\) in a stochastic manner from a set
of parameters \(\thetab \in \mathbb{R}^D\). We denote the process of
obtaining samples from the simulator with
\( m_r(\thetab) \rightarrow \yb \). As shown by~\citet{Meeds2015}, it
is feasible to isolate the randomness of the simulator by introducing
a set of nuisance random variables denoted by \(\ub \sim
p(\ub)\). Therefore, for a specific tuple \((\thetab, \ub)\) the
simulator becomes a deterministic mapping \(g\), such that
\(\yb=\simulator(\thetab,\ub)\). In terms of computer code, the
randomness of a random process is governed by the global seed. There
are some differences on how each scientific package handles the
randomness; for example, at \pkg{Numpy}~\citet{harris2020array} the
pseudo-random number generation is based on a global state, whereas,
at \pkg{JAX}~\citet{jax2018github} random functions consume a key that
is passed as a parameter. However, in all cases, setting the initial
seed to a specific integer converts the simulation to a deterministic
piece of code.

The modeling freedom of simulator-based models comes at the price of
difficulties in inferring the parameters of interest. Denoting the
observed data as \(\data\), the main difficulty lies at the
intractability of the likelihood function
\(L(\thetab) = p(\data|\thetab)\). To better see the sources of the
intractability, and to address them, we go back to the basic
characterization of the likelihood as the (rescaled) probability of
generating data \(\yb\) that is similar to the observed data
\(\data\), using parameters \(\thetab\). Formally, the likelihood
\(L(\thetab)\) is:

\begin{equation}
  \label{eq:likelihood}
  L(\thetab) = \lim_{\epsilon \to 0} c_\epsilon \int_{\yb \in B_{d,\epsilon}(\data)} p(\yb|\thetab)d\yb =
  \lim_{\epsilon \to 0} c_\epsilon \Pr(\simulator(\thetab, \ub) \in \region(\data)  \mid \thetab)
\end{equation}
where \(c_\epsilon\) is a proportionality factor that depends on
\(\epsilon\) and \(\region(\data)\) is an \(\epsilon\) region around
\(\data\) that is defined through a distance function \(d\), i.e., \
\(\region(\data) := \{\yb: d(\yb, \data) \leq \epsilon \}\). In cases
where the output \(\yb\) belongs to a high dimensional space, it is
common to extract summary statistics \(\Phi\) before applying the
distance \(d\). In these cases, the \(\epsilon\)-region is defined as
\(\region(\data) := \{\yb: d(\Phi(\yb), \Phi(\data)) \leq \epsilon
\}\). In our notation, the summary statistics are sometimes omitted
for simplicity. Equation~\ref{eq:likelihood} highlights the main
source of intractability; computing
\(\Pr(\simulator(\thetab,\ub) \in \region(\data) | \thetab) \) as the
fraction of samples that lie inside the \(\epsilon\) region around
\(\data\) is computationally infeasible in the limit where
\(\epsilon \to 0\). Hence, the constraint is relaxed to
\(\epsilon > 0\), which leads to the approximate likelihood:

\begin{equation}
  \label{eq:approx_likelihood}
  L_{d, \epsilon}(\thetab) = \Pr(\yb \in \region(\data) \mid \thetab), \quad \text{where  } \epsilon > 0.
\end{equation}

and, in turn, to the approximate posterior:

\begin{equation} \label{eq:approx_posterior}
  p_{d,\epsilon}(\thetab|\data)
  \propto L_{d, \epsilon}(\thetab) p(\thetab)
\end{equation}

Equations~\ref{eq:approx_likelihood} and~\ref{eq:approx_posterior} is by
no means the only strategy to deal with the intractability of the
likelihood function in Equation~\ref{eq:likelihood}. Other strategies include
modeling the (stochastic) relationship between \(\thetab\) and
\(\yb\), and its reverse, or framing likelihood-free inference as a
ratio estimation problem, see for example \citet{blum2010, Wood2006,
  Papamakarios2016, Papamakarios2019, Chen2019, Thomas2020,
  Hermans2020}. However, both OMC and robust OMC, which we introduce
next, are based on the approximation in Equation~\ref{eq:approx_likelihood}.

\subsection{Optimization Monte Carlo (OMC)}

Our description of OMC~\citet{Meeds2015}
follows~\citet{Ikonomov2019}. We define the indicator function (boxcar
kernel) that equals one only if \(\xb\) lies in \(\region(\yb)\):
\begin{equation}
  \label{eq:indicator}
  \indicator{\region(\yb)}(\xb)=
  \left\{
    \begin{array}{ll}
      1 & \mbox{if } \xb \in \region(\yb) \\
      0 & \mbox{otherwise}
    \end{array} \right. \end{equation}

We, then, rewrite the approximate likelihood function
\(L_{d, \epsilon}(\thetab)\) of Equation~\ref{eq:approx_likelihood} as:

\begin{gather}
  \label{eq:approx_likelihood_omc}
  L_{d, \epsilon}(\thetab) = \Pr(\yb \in \region(\data) | \thetab) =
  \int_{\ub} \indicator{\region(\data)}(g(\thetab, \ub)) d \ub
\end{gather}

which can be approximated using samples from the simulator:

\begin{equation}
  \label{eq:samples_approx_likelihood_omc}
  L_{d, \epsilon}(\thetab) \approx \frac{1}{N} \sum_{i=1}^N \indicator{\region (\data)} (\simulator(\thetab, \ub_i))
 \quad \text{ where } \ub_i \sim p(\ub).
\end{equation}

In Equation~\ref{eq:samples_approx_likelihood_omc}, for each
\(\ub_i\), there is a region \(\accregioni\) in the parameter space
\(\thetab\) where the indicator function returns one, i.e.,
\(\accregioni = \{ \thetab: \simulator(\thetab, \ub_i) \in
\region(\data) \}\). Therefore, we can rewrite the approximate
likelihood and posterior as:

\begin{equation}
  \label{eq:alt_view_theta}
  L_{d, \epsilon}(\thetab) \approx \frac{1}{N} \sum_{i=1}^N \indicator{\accregioni}(\thetab)
\end{equation}

\begin{equation}
  \label{eq:approx_posterior_omc}
  p_{d,\epsilon}(\thetab|\data) \propto
  p(\thetab) \sum_i^N  \indicator{\accregioni}(\thetab).
\end{equation}

As argued by \citet{Ikonomov2019}, these derivations provide a unique
perspective for likelihood-free inference by shifting the focus onto
the geometry of the acceptance regions \(\accregioni\). Indeed, the
task of approximating the likelihood and the posterior becomes a task
of characterizing the sets \(\accregioni\). OMC by \citet{Meeds2015}
assumes that the distance \(d\) is the Euclidean distance
\(||\cdot||_2\) between summary statistics \(\Phi\) of the observed
and generated data, and that the \(\accregioni\) can be well
approximated by infinitesimally small ellipses. These assumptions lead
to an approximation of the posterior in terms of weighted samples
\(\thetab_i^*\) that achieve the smallest distance between observed
and simulated data for each realization \(\ub_i \sim p(\ub)\), i.e.,\
\begin{equation} \label{eq:omc_opt_prob}
\thetab_i^* = \argmin_{\thetab} ||\Phi(\data)-\Phi(\simulator(\thetab, \ub_i))||_2  , \quad \ub_i \sim p(\ub).
\end{equation}
The weighting for each \(\thetab_i^*\) is proportional to the prior
density at \(\thetab_i^*\) and inversely proportional to the determinant
of the Jacobian matrix of the summary statistics at \(\thetab_i^*\). For
further details on OMC we refer the reader to \citet{Meeds2015,
  Ikonomov2019}.

\subsection{Robust optimization Monte Carlo (ROMC)}

\citet{Ikonomov2019} showed that considering infinitesimally small
ellipses can lead to highly overconfident posteriors. We refer the
reader to their paper for the technical details and conditions for
this issue to occur. Intuitively, it happens because the weights in
OMC are only computed from information at \(\thetab_i^*\), and using
only local information can be misleading. For example, if the
curvature of \(||\Phi(\data)-\Phi(\simulator(\thetab, \ub_i))||_2\) at
\(\thetab_i^*\) is nearly flat, it may wrongly indicate that
\(\accregioni\) is much larger than it actually is. In our software
package we implement the robust generalization of OMC by
\citet{Ikonomov2019} that resolves this issue.

ROMC approximates the acceptance regions \(\accregioni\) with
\(D\)-dimensional bounding boxes \(\accregionihat\). A uniform
distribution, \(q_i(\thetab)\), is defined on top of each bounding box
and serves as a proposal distribution for generating posterior samples
\(\thetab_{ij} \sim q_i\). The samples get an (importance) weight
\(w_{ij}\) that compensate for using the proposal distributions
\(q_i(\thetab)\) instead of the prior \(p(\thetab)\):

\begin{equation}
  \label{eq:sampling}
  w_{ij} = \indicator{\accregioni}(\thetab_{ij}) \frac{ p(\thetab_{ij})}{q(\thetab_{ij})}.
\end{equation}

Given the weighted samples, any expectation
\(\E_{p(\thetab|\data)}[h(\thetab)]\) of some function \(h(\thetab)\), can be approximated as
\begin{equation} \label{eq:expectation}
  \E_{p(\thetab|\data)}[h(\thetab)] \approx \frac{\sum_{ij} w_{ij} h(\thetab_{ij})}{\sum_{ij} w_{ij}}
\end{equation}

The approximation of the acceptance regions contains two
compulsory and one optional step: (i) solving the optimization
problems as in OMC, (ii) constructing bounding boxes around
\(\accregioni\) and, optionally, (iii) refining the approximation via a
surrogate model of the distance.

\subsubsection*{(i) Solving the deterministic optimization problems}
For each set of nuisance variables \(\ub_i, i = \{1,2,\ldots,n_1 \}\),
we search for a point \(\thetab^*_i\) such that
\(d(\simulator(\thetab^*_i,\ub_i), \data) \le \epsilon\). In
principle, \(d(\cdot)\) can refer to any valid distance function. For
the rest of the paper we consider \(d(\cdot)\) as the squared
Euclidean distance, as in~\citet{Ikonomov2019}. For simplicity, we use
\(d_i(\thetab)\) to refer to \(d(\simulator(\thetab,\ub_i),
\data)\). We search for \(\theta_i^*\) solving:

\begin{equation}
  \label{eq:optProb}
  \thetab_i^* = \!\argmin_{\thetab} d_i(\thetab)
\end{equation}
and we accept the solution only if it satisfies the constraint
\(d_i(\thetab_i^*) \le \epsilon\). If \(d_i(\thetab)\) is
differentiable, Equation~\ref{eq:optProb} can be solved using any
gradient-based optimizer. The gradients
\(\nabla_{\thetab} d_i(\thetab)\) can be either provided in closed
form or approximated by finite differences. If \(d_i\) is not
differentiable, Bayesian Optimization~\citep{Shahriari2016} can be
used instead. In this scenario, apart from obtaining an optimal
\(\thetab_i^* \), we can also automatically build a surrogate model
\(\hat{d}_i(\thetab)\) of the distance function \(d_i(\thetab)\). The
surrogate model \(\hat{d}_i\) can then substitute the actual distance
function in downstream steps of the algorithms, with possible
computational gains especially in cases where evaluating the actual
distance \(d_i(\thetab)\) is expensive.

\subsubsection*{(ii) Estimating the acceptance regions}

Each acceptance region \(\accregioni\) is approximated by a bounding
box \(\accregionihat\). The acceptance regions \(\accregioni\) can
contain any number of disjoint subsets in the \(D\)-dimensional space
and any of these subsets can take any arbitrary shape. We should make
three important remarks. First, since the bounding boxes are built
around \(\thetab_i^*\), we focus only on the connected subset of
\(\accregioni\) that contains \(\thetab_i^*\), which we denote as
\(\accregionitheta\). Second, we want to ensure that the bounding box
\(\accregionihat\) is big enough to contain on its interior all the
volume of \(\accregionitheta\). Third, we want \(\accregionihat\) to
be as tight as possible to \(\accregionitheta\) to ensure high
acceptance rate on the importance sampling step that
follows. Therefore, the bounding boxes are built in two steps. First,
we compute their axes \(\mathbf{v}_m\), for \(m = \{1, \ldots, D\}\)
based on the (estimated) curvature of the distance at \(\thetab_i^*\),
and, second, we apply a line-search method along each axis to
determine the size of the bounding box. We refer the reader to
Algorithm~\ref{alg:region_construction} for the details. After the
bounding boxes construction, a uniform distribution \(q_i\) is defined
on each bounding box, and is used as the proposal region for
importance sampling.

\subsubsection*{(iii) Refining the estimate via a local surrogate model (optional)}

For computing the weight \(w_{ij}\) at Equation~\ref{eq:sampling}, we
need to check whether the samples \(\thetab_{ij}\), drawn from the
bounding boxes, are inside the acceptance region \(\accregioni\). This
can be considered to be a safety-mechanism that corrects for any
inaccuracies in the construction of \(\accregionihat\) above. However,
this check involves evaluating the distance function
\(d_i(\thetab_{ij})\), which can be expensive if the model is
complex. \citet{Ikonomov2019} proposed fitting a surrogate model
\(\tilde{d}_i(\thetab)\) of the distance function \(d_i(\thetab)\), on
data points that lie inside \(\accregionihat\). In principle, any
regression model can be used as surrogate model. \citet{Ikonomov2019}
used a simple quadratic model because it has ellipsoidal isocontours,
which facilitates replacing the bounding box approximation of
\(\accregioni\) with a tighter-fitting ellipsoidal approximation.

The training data for the quadratic model is obtained by sampling
\(\thetab_{ij} \sim q_i\) and accessing the distances
\(d_i(\thetab_{ij})\). The generation of the training data adds an
extra computational cost, but leads to a significant speed-up when
evaluating the weights \(w_{ij}\). Moreover, the extra cost is largely
eliminated if Bayesian optimization with a Gaussian process (GP)
surrogate model \(\hat{d}_i(\thetab)\) was used to obtain
\(\thetab_i^*\) in the first step. In this case, we can use
\(\hat{d}_i(\thetab)\) instead of \(d_i(\thetab)\) to generate the
training data. This essentially replaces the global GP model with a
simpler local quadratic model which is typically more robust.

\subsection[Engine for likelihood-free inference (ELFI)]{\pkg{Engine for likelihood-free inference (ELFI)}}
\label{subsec:ELFI}

\begin{figure}[ht]
    \begin{center}
      \includegraphics[width=0.8\textwidth]{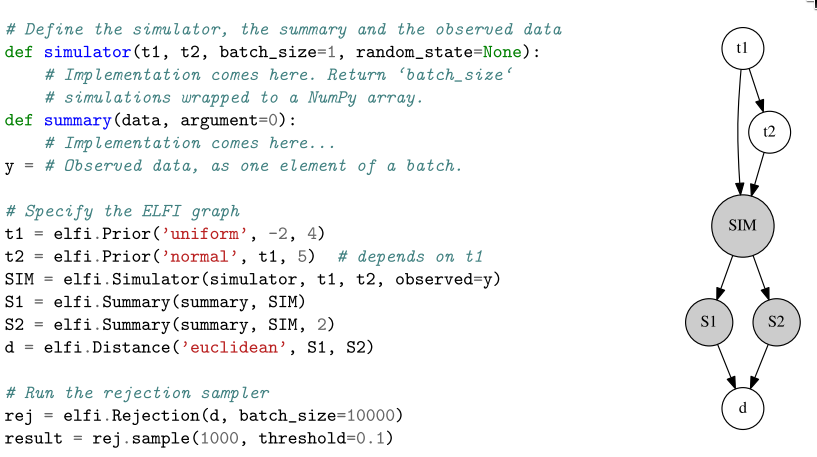}
    \end{center}
    \caption[Baseline example for creating an \pkg{ELFI} model]{Baseline example for creating an \pkg{ELFI} model. Image taken from \cite{1708.00707}}
    \label{fig:elfi}
\end{figure}

\pkg{Engine for Likelihood-Free Inference (ELFI)}\footnote{Extended
  documentation can be found
  \href{https://elfi.readthedocs.io/en/latest/}{https://elfi.readthedocs.io}}~\cite{1708.00707}
is a \proglang{Python} package for LFI. We selected to implemented
ROMC in \pkg{ELFI} since it provides convenient modules for all the
fundamental components of a probabilistic model (e.g.\ prior,
simulator, summaries etc.). Furthermore, \pkg{ELFI} already supports
some recently proposed likelihood-free inference methods. \pkg{ELFI}
handles the probabilistic model as a Directed Acyclic Graph
(DAG). This functionality is based on the package
\pkg{NetworkX}~\cite{hagberg2008exploring}, which supports
general-purpose graphs. In most cases, the structure of a
likelihood-free model follows the pattern of Figure~\ref{fig:elfi};
some edges connect the prior distributions to the simulator, the
simulator is connected to the summary statistics that, in turn, lead
to the output node. Samples can be obtained from all nodes through
sequential (ancestral) sampling. \pkg{ELFI} automatically considers as
parameters of interest, i.e., those we try to infer a posterior
distribution, the ones included in the \code{elfi.Prior} class.

All inference methods of \pkg{ELFI} are implemented following two
conventions. First, their constructor follows the signature
\code{elfi.<Class name>(<output node>, *arg)}, where \code{<output
  node>} is the output node of the simulator-based model and
\code{*arg} are the parameters of the method. Second, they provide a
method \code{elfi.<Class name>.sample(*args)} for drawing samples from
the approximate posterior.

%% -- Design Principles -----------------------------------------------------
\section{Overview of the implementation}
\label{sec:implementation}
In this section, we express ROMC as an algorithm and then we present
the general implementation principles.

\subsection{Algorithmic view of ROMC}

For designing an extendable implementation, we firstly define ROMC as
a sequence of algorithmic steps. At a high level, ROMC can be split
into the training and the inference part; the training part covers the
steps for estimating the proposal regions and the inference part
calculates the weighted samples. In
Algorithm~\ref{alg:romc_algorithm}, that defines ROMC as an algorithm,
steps 2-11 (before the horizontal line) refer to the training part and
steps 13-18 to the inference part.

\begin{algorithm}[!ht]
  \caption{ROMC. Requires the prior \( p(\thetab) \), the simulator
    \(M_r(\thetab)\), number of optimization problems \(n_1\), number
    of samples per region \(n_2\), acceptance limit
    \(\epsilon\)}\label{alg:romc_algorithm}
  \begin{algorithmic}[1]
    \Procedure{ROMC}{}
    \For{\(i \gets 1 \textrm{ to } n_1\)}
    \State \(\ub_i \sim p(\ub)\) \Comment{Draw nuisance variables}
    \State Convert \(M_r(\thetab) \) to \( g(\thetab, \ub=\ub_i) \) \Comment{Define deterministic simulator}
      \State \( d_i(\thetab) = d(g(\thetab, \ub=\ub_i), \yb_0) \) \Comment{Define distance function}
      \State \(\thetab_i^* = \text{argmin}_{\thetab} d_i\), \(d_i^*=d_i(\thetab_i^*)\) \Comment{Solve optimization problem}
      \If{\(d^*_i > \epsilon\)}
        \State Go to 2 \Comment{Filter solution}
      \EndIf
      \State Estimate \(\hat{\accregioni}\) and define \(q_i\) \Comment{Estimate proposal area}
      \State (Optional) Fit \(\Tilde{d}_i\) on \(\hat{\accregioni}\) \Comment{Fit surrogate model}
      \\\hrulefill
      \For{\(j \gets 1 \textrm{ to } n_2\)}
      \State \(\thetab_{ij} \sim q_i\), compute \(w_{ij}\) as in Algorithm \ref{alg:sampling_GB} \Comment{Sample}
      \EndFor
    \EndFor
    \State \(\E_{p(\thetab|\data)}[h(\thetab)]\) as in eq.~\eqref{eq:expectation} \Comment{Estimate an expectation}
    \State \(p_{d,\epsilon}(\thetab) \) as in eq.~\eqref{eq:approx_posterior_omc} \Comment{Evaluate the unnormalized posterior}
    \EndProcedure
  \end{algorithmic}
\end{algorithm}

\subsubsection*{Training part}
\noindent
At the training (fitting) part, the goal is the estimation of the
proposal regions \(\hat{\accregioni}\), which expands to three
mandatory tasks; (a) sample the nuisance variables
\(\ub_i \sim p(\ub)\) for defining the deterministic distance
functions \(d_i(\thetab)\) (steps 3-5), (b) solve the optimization
problems for obtaining \(\thetab_i^*, d_i^*\) and keep the solutions
inside the threshold \(\epsilon\) (steps 6-9), and (c) estimate the
bounding boxes \(\accregionihat\) to define uniform distributions
\(q_i\) on them (step 10). Optionally, a surrogate model
\(\tilde{d}_i\) can be fitted for a faster inference phase (step 11).

If \(d_i(\thetab)\) is differentiable, using a gradient-based method
is advised for obtaining \(\thetab_i^*\) faster. In this case, the
gradients \(\nabla_{\thetab} d_i\) gradients are approximated
automatically with finite-differences, if they are not provided in
closed-form by the user. Finite-differences approximation requires two
evaluations of \(d_i\) for each parameter
\(\theta_m, m \in \{1, \ldots, D\}\), which scales well only in
low-dimensional problems. If \(d_i(\thetab)\) is not differentiable,
Bayesian Optimization can be used as an alternative. In this scenario,
the training part becomes slower due to fitting of the surrogate model
and the blind optimization steps.

After obtaining the optimal points \(\thetab^*_i\), we estimate the
proposal regions. Algorithm~\ref{alg:region_construction} describes
the line search approach for finding the region boundaries. The axes
of each bounding box \(\vb_m, m = \{1, \ldots, D\}\) are defined as
the directions with the highest curvature at \(\thetab_i^*\) computed
by the eigenvalues of the Hessian matrix \(\hessian_i\) of \(d_i\) at
\(\thetab_i\) (step 1). Depending on the algorithm used in the
optimization step, we either use the real distance \(d_i\) or the
Gaussian Process approximation \(\hat{d}_i\). When the distance
function is the Euclidean distance (default choice), the Hessian
matrix can be also computed as \(\hessian_i = \jac_i^T\jac_i\), where
\(\jac_i\) is the Jacobian matrix of the summary statistics
\(\Phi(g(\thetab, \ub_i))\) at \(\thetab_i^*\). This approximation has
the computational advantage of using only first-order
derivatives. After defining the axes, we search for the bounding box
limits with a line step algorithm (steps 2-21). The key idea is to
take long steps \(\eta\_\text{start}\) until crossing the boundary and
then take small steps backwards to find the exact boundary position.

\begin{algorithm}[!ht]
  \caption{Approximation \(\accregioni\) with a bounding box \(\hat{\accregioni}\);
  Requires: a model of distance \(d_i(\thetab)\),
  an optimal point \(\thetab_i^*\),
  a number of refinements \(K\),
  a step size \(\eta\_\text{start}\),
  maximum iterations \(M\) and
  a curvature matrix \(\hessian_i\) (\(\jac_i^T\jac_i \) or GP Hessian)}\label{alg:region_construction}
  \begin{algorithmic}[1]
  \State Compute eigenvectors \(\vb_{m}\) of \(\hess_i\) {\scriptsize (\(m = 1,\ldots,D)\)}
  \For{\(m \gets 1 \textrm{ to } D\)}
    \State \(\Tilde{\thetab} \gets \thetab_i^*\) \label{algstep:box_constr_start}
    \State \(k \gets 0\)
    \State \(\eta \gets \eta\_\text{start}\) \Comment{Initialize \(\eta\)}
    \Repeat
          \State \(j \gets 0\)
          \Repeat
            \State \(\Tilde{\thetab} \gets \Tilde{\thetab} + \eta \ \vb_{m}\) \Comment{Large step size \(\eta\).}
            \State \(j \gets j + 1\)
          \Until{\(d(g(\Tilde{\thetab}, \ub=\ub_i), \data) > \epsilon\) or \(j \geq M\)} \Comment{Check distance or maximum iterations}
          \State \(\Tilde{\thetab} \gets \Tilde{\thetab} - \eta \ \vb_{m}\)
          \State \(\eta \gets \eta/2\) \Comment{More accurate region boundary}
          \State \(k \gets k + 1\)
      \Until \(k = K\)
      \If{\(\Tilde{\thetab} = \thetab_i^*\)} \Comment{Check if no step has been done}
        \State \(\Tilde{\thetab} \gets \Tilde{\thetab} + \dfrac{\eta\_{\text{start}}}{2^K} \vb_m\) \Comment{Then, make the minimum step}
      \EndIf \label{algstep:box_constr_end}
      \State Set \(\Tilde{\thetab}\) as the positive end point along \(\vb_m\)
      \State Run steps~\ref{algstep:box_constr_start}~-~\ref{algstep:box_constr_end} for \(\vb_m = - \vb_m\) and set \(\Tilde{\thetab}\) as the negative end point along \(\mathbf{v}_{m}\)
  \EndFor
  \State Fit a rectangular box around the region end points and define \(q_i\) as uniform distribution
  \end{algorithmic}
\end{algorithm}

\subsubsection*{Inference Part}

The inference part includes one or more of the following three tasks;
(a) sample from the posterior distribution
\( \thetab_i \sim p_{d, \epsilon}(\thetab|\data)\)
(Equation~\ref{eq:sampling}), (b) compute an expectation
\(\E_{\thetab|\data}[h(\thetab)]\) (Equation~\ref{eq:expectation})
and/or (c) evaluate the unnormalized posterior
\(p_{d, \epsilon}(\thetab|\data)\)
(Equation~\ref{eq:approx_posterior_omc}). Sampling is performed by
getting \(n_2\) samples from each proposal distribution \(q_i\). For
each sample \(\thetab_{ij}\), the distance function\footnote{As
  before, a surrogate model \(\hat{d}\) can be utilized as the
  distance function if it is available.} is evaluated for checking if
it lies inside the acceptance region. Algorithm~\ref{alg:sampling_GB}
defines the steps for computing a weighted sample. After we obtain
weighted samples, computing the expectation is straightforward using
Equation~\ref{eq:expectation}. Finally, evaluating the unnormalized
posterior at a specific point \(\thetab\) requires access to the
distance functions \(d_i\) and the prior distribution
\(p(\thetab)\). Following Equation~\ref{eq:approx_posterior_omc}, we
simply count for how many deterministic distance functions it holds
that \(d_i(\thetab) < \epsilon\). It is worth noticing that for
evaluating the unnormalized posterior, there is no need for solving
the optimization problems and building the proposal regions.

\begin{algorithm}[H]
    \centering
    \caption{Sampling. Requires a function of distance \(d_i\), the prior distribution \(p(\thetab)\), the proposal distribution \(q_i\)}\label{alg:sampling_GB}
    \begin{algorithmic}[1]
      \State \(\thetab_{ij} \sim q_i \forall i\) \Comment{Sample parameters}
      \For {\(i \gets 1 \textrm{ to } n_1\)}
      \For {\(j \gets 1 \textrm{ to } n_2\)}
          \If {\(d_i(\thetab_{ij}) \leq \epsilon\)} \Comment{Accept sample}
            \State \(w_{ij} = \frac{p(\thetab_{ij})}{q(\thetab_{ij})}\) \Comment{Compute weight}
            \State Store \((w_{ij}, \thetab_{ij})\) \Comment{Store weighted sample}
            \EndIf
            \EndFor
    \EndFor
    \end{algorithmic}
\end{algorithm}

\subsection{General implementation principles}
\label{subsec:general_design}

\tikzstyle{startstop} = [rectangle, rounded corners, minimum width=3cm, minimum height=1cm,text centered, draw=black, fill=red!30]
\tikzstyle{io} = [trapezium, trapezium left angle=70, trapezium right angle=110, minimum width=3cm, minimum height=1cm, text centered, draw=black, fill=blue!30]
\tikzstyle{train_process} = [rectangle, minimum width=3cm, minimum height=.7cm, text centered, draw=black, fill=green!40]
\tikzstyle{infer_process} = [rectangle, minimum width=3cm, minimum height=.7cm, text centered, draw=black, fill=blue!30]

\tikzstyle{decision} = [diamond, minimum width=.1cm, minimum height=.1cm, text centered, draw=black, fill=green!30]

\tikzstyle{public_func_rev} = [draw=black, rotate=90, anchor=north, fill=blue!20, rounded corners]
\tikzstyle{public_func} = [draw=black, fill=blue!20, rounded corners]
\tikzstyle{arrow} = [thick,->,>=stealth]

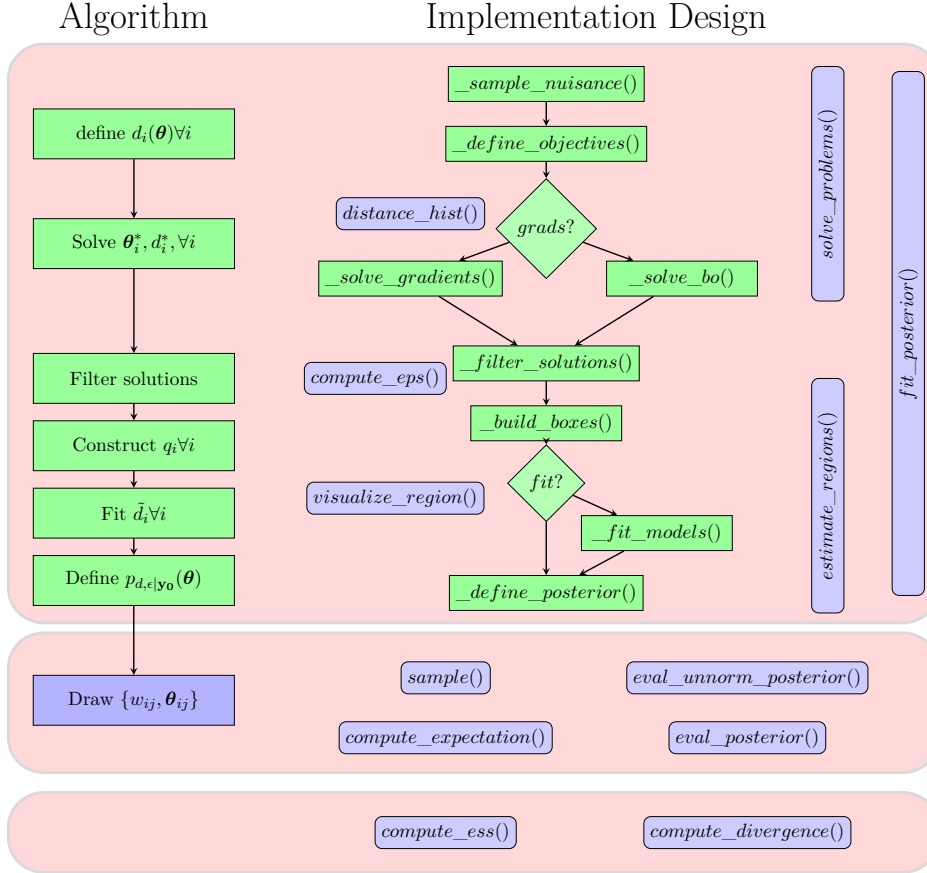
\begin{figure}[ht]
  \begin{center}
  %   \resizebox{.32\textwidth}{!}{
  %     \begin{tikzpicture}[node distance=1.4cm, scale=.1]
  %     \end{tikzpicture}
  %   }
    \resizebox{.8\textwidth}{!}{    
      \begin{tikzpicture}[node distance=1.2cm, scale=.1]

        % backgrounds
        \draw [ultra thick, draw=black, fill=red, opacity=0.15, rounded corners=20pt] (-115, 8) rectangle (70, -108);
        \draw [ultra thick, draw=black, fill=red, opacity=0.15, rounded corners=20pt] (-115, -110) rectangle (70, -138);
        \draw [ultra thick, draw=black, fill=red, opacity=0.15, rounded corners=20pt] (-115, -142) rectangle (70, -158);

        % first graph
        % private functions
        \node (n1) [train_process, xshift=-.8cm] { $\_sample\_nuisance()$  };
        \node (n2) [train_process, below of=n1] { $\_define\_objectives()$  };
        \node (n3) [decision, below of=n2, yshift=-.5cm] { $grads?$  };
        \node (n4) [train_process, left of=n3, yshift=-1cm, xshift=-1.5cm] { $\_solve\_gradients()$  };
        \node (n5) [train_process, right of=n3, yshift=-1cm, xshift=1.5cm] { $\_solve\_bo()$  };
        \node (n6) [train_process, below of=n3, yshift=-1.5cm] { $\_filter\_solutions()$  };
        \node (n7) [train_process, below of=n6] { $\_build\_boxes()$  };
        \node (n8) [decision, below of=n7] { $fit?$  };
        \node (n9) [train_process, right of=n8, yshift=-1cm, xshift=1cm] { $\_fit\_models()$  };

        \node (n10) [train_process, below of=n8, yshift=-1cm] { $\_define\_posterior()$  };

        % public functions
        \node (n11) [public_func_rev, right of=n3, yshift=-5.6cm, xshift=-0.3cm, minimum width=4.7cm] {$solve\_problems()$};
        \node (n12) [public_func_rev, right of=n9, yshift=-3.4cm, xshift=-.45cm, minimum width=4.7cm] {$estimate\_regions()$};
        \node (n13) [public_func_rev, right of=n5, yshift=-4.5cm, xshift=-2.3cm, minimum width=10.5cm] {$fit\_posterior()$};
        \node (n14) [public_func, left of=n3, yshift=0.3cm, xshift=-1.5cm] {$distance\_hist()$};
        \node (n15) [public_func, left of=n6, xshift=-2.2cm, yshift=-.3cm] {$compute\_eps()$};        
        \node (n16) [public_func, left of=n7, xshift=-1.8cm, yshift=-1.5cm] {$visualize\_region()$};

        % public functions inference
        \node (n17) [public_func, below of=n10, xshift=-2cm, yshift=-.5cm] {$sample()$};
        \node (n18) [public_func, below of=n17] {$compute\_expectation()$};        
        \node (n19) [public_func, below of=n10, xshift=4cm, yshift=-.5cm] {$eval\_unnorm\_posterior()$};
        \node (n20) [public_func, below of=n19] {$eval\_posterior()$};

        % public functions evaluation
        \node (n21) [public_func, below of=n18, yshift=-.7cm] {$compute\_ess()$};
        \node (n22) [public_func, below of=n20, yshift=-.7cm] {$compute\_divergence()$};        

        % add headers
        \node (impl_design) [below of=n1, yshift=2.5cm, xshift=1cm, minimum width=4cm, minimum height=1cm] {\huge{Implementation Design} };

        % arrows
        \draw [arrow] (n1) -- (n2);
        \draw [arrow] (n2) -- (n3);
        \draw [arrow] (n3) -- (n4);
        \draw [arrow] (n3) -- (n5);
        \draw [arrow] (n4) -- (n6);
        \draw [arrow] (n5) -- (n6);
        \draw [arrow] (n6) -- (n7);
        \draw [arrow] (n7) -- (n8);
        \draw [arrow] (n8) -- (n9);
        \draw [arrow] (n9) -- (n10);
        \draw [arrow] (n8) -- (n10);

        % second graph
        \node (pro1) [train_process, left of=n1, xshift=-7cm, yshift=-1cm, minimum width=4cm, minimum height=1cm] { define $d_i(\thetab) \forall i$  };
        \node (pro2) [train_process, below of=pro1, yshift=-1cm, minimum width=4cm, minimum height=1cm] { Solve $\thetab_i^*, d_i^*, \forall i$  };
        \node (filter) [train_process, below of=pro2, yshift=-1.5cm, minimum width=4cm, minimum height=1cm] { Filter solutions  };
        \node (proposal_region) [train_process, below of=filter, yshift=-0.15cm, minimum width=4cm, minimum height=1cm] {Construct $q_i \forall i$};
        \node (surrogate) [train_process, below of=proposal_region, yshift=-0.15cm, minimum width=4cm, minimum height=1cm] {Fit $\Tilde{d}_i \forall i$};
        \node (posterior) [train_process, below of=surrogate, yshift=-0.15cm, minimum width=4cm, minimum height=1cm] {Define $p_{d,\epsilon|\data}(\thetab)$};
        
        \node (sample) [infer_process, below of=posterior, yshift=-1.2cm, minimum width=4cm, minimum height=1cm] {Draw $\{w_{ij}, \thetab_{ij} \}$ };

        % add headers
        \node (algorithm) [below of=pro1, yshift=3.5cm, minimum width=4cm, minimum height=1cm] {\huge{Algorithm} };
        
        \draw [arrow] (pro1) -- (pro2);
        \draw [arrow] (pro2) -- (filter);
        \draw [arrow] (filter) -- (proposal_region);
        \draw [arrow] (proposal_region) -- (surrogate);
        \draw [arrow] (surrogate) -- (posterior);
        \draw [arrow] (posterior) -- (sample);

      \end{tikzpicture}
    }
  \end{center}
  \caption{Overview of the ROMC implementation. On the left side, we
    depict ROMC as a sequence of algotirhmic steps. On the right side,
    we present the functions that form our implementation; the green
    rectangles (starting with underscore) are the internal
    functionalities and the blue rectangles the publicly exposed
    API. This side-by-side illustration highlights that our
    implementation follows strictly the algorithmic view of ROMC.}
\label{fig:romc_overview}
\end{figure}

The overview of our implementation is illustrated in
Figure~\ref{fig:romc_overview}. Following \proglang{Python} naming
principles, the methods starting with an underscore (green rectangles)
represent internal (private) functions, whereas the rest (blue
rectangles) are the methods exposed at the API. In
Figure~\ref{fig:romc_overview}, it can be observed that the
implementation follows Algorithm~\ref{alg:romc_algorithm}. The
training part includes all the steps until the computation of the
proposal regions, i.e., sampling the nuisance variables, defining the
optimization problems, solving them, constructing the regions and
fitting local surrogate models. The inference part comprises of
evaluating the unnormalized posterior (and the normalized when is
possible), sampling and computing an expectation. We also provide some
utilities for inspecting the training process, such as plotting the
histogram of the final distances or visualizing the constructed
bounding boxes. Finally, in the evaluation part, we provide two
methods for evaluating the inference; (a) computing the Effective
Sample Size (ESS) of the samples and (b) measuring the divergence
between the approximate posterior the ground-truth, if the latter is
available.\footnote{Normally, the ground-truth posterior is not
  available; However, this functionality is useful in cases where the
  posterior can be computed numerically or with an alternative method,
  e.g., Rejection Sampling, and we want to measure the discrepancy
  between the two approximations.}

\subsubsection*{Parallel version of ROMC}

As discussed, ROMC has the significant advantage of being fully
parallelisable. We exploit this fact by implementing a parallel
version of the major fitting components; (a) solving the optimization
problems, (b) constructing bounding box regions. We parallelize these
processes using the built-in \proglang{Python} package
\pkg{multiprocessing}. The specific package enables concurrency, using
sub-processes instead of threads, for side-stepping the Global
Interpreter (GIL). For activating the parallel version of the
algorithm, the user simply has to set \\
\code{elfi.ROMC(<output_node>, parallelize = True)}.

\subsubsection*{Simple one-dimensional example}

For illustrating the functionalities we will use the following running
example introduced by \citet{Ikonomov2019},

\begin{gather} \label{eq:1D_example}
  p(\theta) = \mathcal{U}(\theta;-2.5,2.5)\\ \label{eq:1D_example_eq_2}
  p(y|\theta) = \left\{
    \begin{array}{ll} \theta^4 + u & \mbox{if } \theta \in [-0.5, 0.5]
\\ |\theta| - c + u & \mbox{otherwise}
    \end{array} \right.\\
  u \sim \mathcal{N}(0,1)
\end{gather}

\noindent

The prior is a uniform distribution in the range \([-2.5, 2.5]\) and
the likelihood is defined at Equation~\ref{eq:1D_example_eq_2}. The
constant \(c\) is \(0.5 - 0.5^4\) ensures that the PDF is
continuous. There is only one observation \(y_0 = 0\). The inference
in this particular example can be performed quite easily without using
a likelihood-free inference approach. We can exploit this fact for
validating the accuracy of our implementation. At the following code
snippet, we code the model at \pkg{ELFI}:

\begin{Code}
import elfi
import scipy.stats as ss
import numpy as np

def simulator(t1, batch_size = 1, random_state = None):
    c = 0.5 - 0.5**4
    if t1 < -0.5:
        y = ss.norm(loc = -t1-c, scale = 1).rvs(random_state = random_state)
    elif t1 <= 0.5:
        y = ss.norm(loc = t1**4, scale = 1).rvs(random_state = random_state)
    else:
        y = ss.norm(loc = t1-c, scale = 1).rvs(random_state = random_state)
    return y

# observation
y = 0

# Elfi graph
t1 = elfi.Prior('uniform', -2.5, 5)
sim = elfi.Simulator(simulator, t1, observed = y)
d = elfi.Distance('euclidean', sim)

# Initialize the ROMC inference method
bounds = [(-2.5, 2.5)] # bounds of the prior
parallelize = False # True activates parallel execution
romc = elfi.ROMC(d, bounds = bounds, parallelize = parallelize)
\end{Code}

%% -- ROMC implementation --
\section{Implemented functionalities}

At this section, we analyze the functionalities of the training, the
inference and the evaluation part. Extended documentation for each
method can be found in \pkg{ELFI}'s
\href{https://elfi.readthedocs.io/en/latest/}{official
  documentation}. Finally, we describe how a user may extend ROMC with
its custom modules.

\subsection{Training part}
\label{subsec:training}

In this section, we describe the six functions of the training part:

\begin{Code}
>>> romc.solve_problems(n1, use_bo = False, optimizer_args = None)
\end{Code}

\noindent
This method (a) draws integers for setting the seed, (b) defines the
optimization problems and (c) solves them using either a
gradient-based optimizer (default choice) or Bayesian optimization
(BO), if \code{use_bo = True}. The tasks are completed sequentially,
as shown in Figure~\ref{fig:romc_overview}. The optimization problems
are defined after drawing \code{n1} integer numbers from a discrete
uniform distribution \(u_i \sim \mathcal{U}\{1, 2^{32}-1\}\), where
each integer \(u_i\) is the seed passed to \pkg{ELFI}'s random
simulator. The user can pass a \code{Dict} with custom parameters to
the optimizer through \code{optimizer_args}. For example, in the
gradient-based case, the user may pass \code{optimizer_args =
  \{"method": "L-BFGS-B", "jac": jac\}}, to select the
\code{"L-BFGS-B"} optimizer and use the callable \code{jac} to compute
the gradients in closed-form.

\begin{Code}
>>> romc.distance_hist(**kwargs)
\end{Code}

\noindent
This function helps the user decide which threshold \(\epsilon\) to
use by plotting a histogram of the distances at the optimal point
\(d_i(\thetab_i^*) : \{i = 1, 2, \ldots, n_1\}\) or \(\hat{d}_i^*\) in
case \code{use_bo = True}. The function forwards the keyword arguments
to the underlying \code{pyplot.hist()} of the \pkg{matplotlib}
package. In this way the user may customize some properties of the
histogram, e.g., the number of bins.

\begin{Code}
>>> romc.estimate_regions(eps_filter, use_surrogate = None, fit_models = False)
\end{Code}

\noindent
This method estimates the proposal regions around the optimal points,
following Algorithm~\ref{alg:region_construction}. The choice about
the distance function follows the previous optimization step; if a
gradient-based optimizer has been used, then estimating the proposal
region is based on the real distance \(d_i\). If BO is used, then the
surrogate distance \(\hat{d}\) is chosen. Setting
\code{use_surrogate=False} enforces the use of the real distance \(d\)
even after BO. Finally, the parameter \code{fit_models} selects
whether to fit local surrogate models \(\tilde{d}\) after estimating
the proposal regions.

\noindent
The training part includes three more functions. The function
\code{romc.fit_posterior(args*)} which is a syntactic sugar for
applying \code{.solve_problems()} and \code{.estimate_regions()} in a
single step. The function \code{romc.visualize_region(i)} plots the
bounding box around the optimal point of the \(i\)-th optimization
problem, when the parameter space is up to \(2D\). Finally,
\code{romc.compute_eps(quantile)} returns the appropriate distance
value \(d_{i=\kappa}^*\) where
\(\kappa = \lfloor quantile \cdot n \rfloor\) from the collection
\(\{ d_i^* \}, i = \{1, \ldots, n\}\) where \(n\) is the number
of accepted solutions. It can be used to automate the selection of the
threshold \(\epsilon\), e.g.,
\code{eps=romc.compute_eps(quantile=0.9)}.

\subsubsection*{Example - Training part}

In the following snippet, we put together the routines to code the
training part of the running example.

\begin{Code}
# Training (fitting) part
n1 = 500 # number of optimization problems
seed = 21 # seed for solving the optimization problems
use_bo = False # set to True for switching to Bayesian optimization

# Training step-by-step
romc.solve_problems(n1 = n1, seed = seed, use_bo = use_bo)
romc.theta_hist(bins = 100) # plot hist to decide which eps to use

eps = .75 # threshold for the bounding box
romc.estimate_regions(eps = eps) # build the bounding boxes

romc.visualize_region(i = 2) # for inspecting visually the bounding box

# Equivalent one-line command
# romc.fit_posterior(n1 = n1, eps = eps, use_bo = use_bo, seed = seed)
\end{Code}

\begin{figure}[ht]
  \begin{center}
    \includegraphics[width=0.49\textwidth]{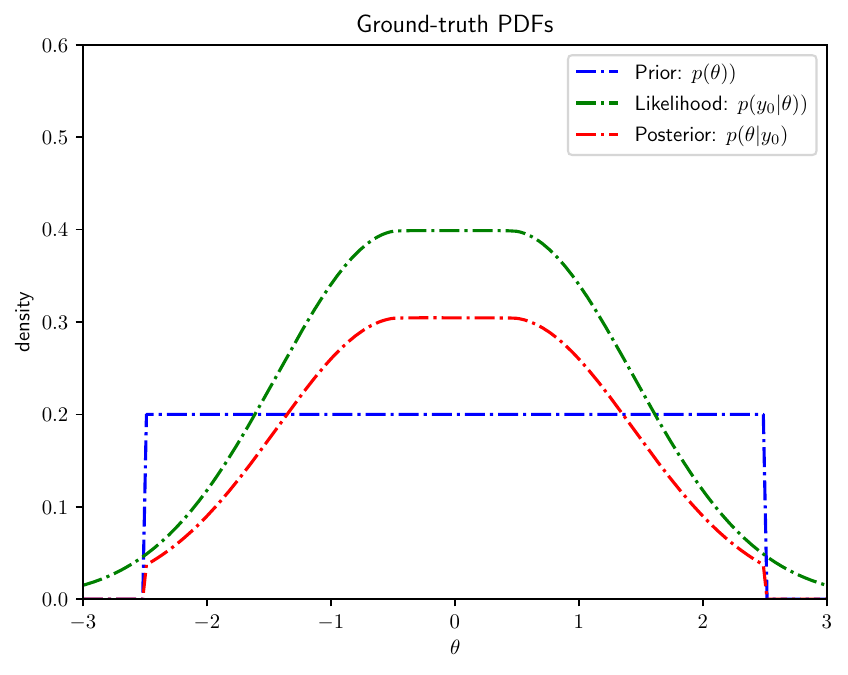}
    \includegraphics[width=0.49\textwidth]{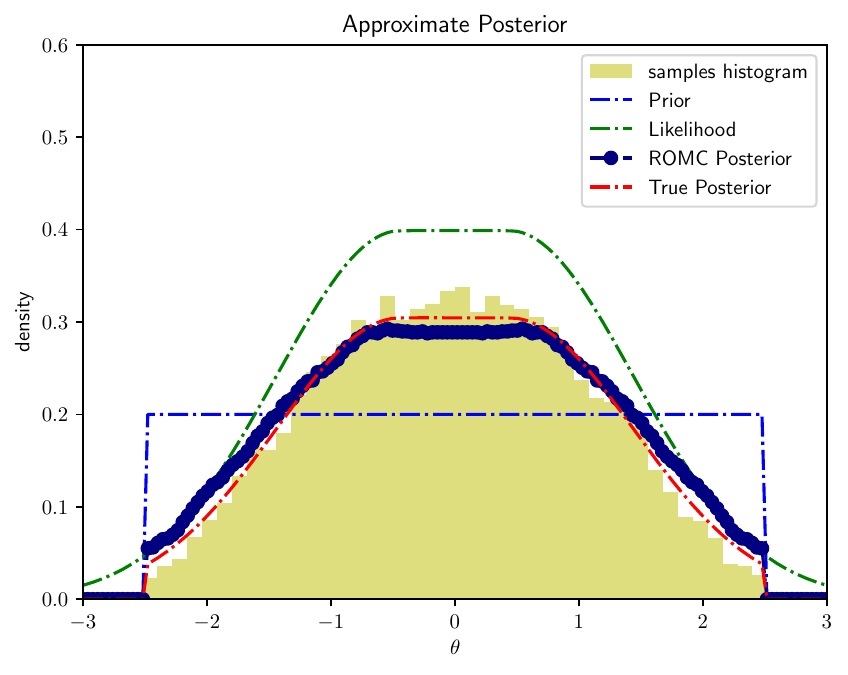}
  \end{center}
  \caption[Histogram of distances at the one-dimensional example.]{Histogram of
    distances and visualization of a specific region.}
  \label{fig:running_example_romc_inference}
\end{figure}

\subsection{Inference part}
\label{subsec:inference}

In this section, we describe the four functions of the inference part:

\begin{Code}
>>> romc.sample(n2)
\end{Code}

\noindent
This is the basic functionality of the inference, drawing \(n_2\)
samples for each bounding box region, giving a total of
\(k \cdot n_2\) samples, where \(k < n_1\) is the number of the
optimal points remained after filtering. The samples are drawn from a
uniform distribution \(q_i\) defined over the corresponding bounding
box and the weight \(w_i\) is computed as in
Algorithm~\ref{alg:sampling_GB}.

\noindent
The inference part includes three more function. The function
\code{romc.compute_expectation(h)} computes the expectation
\(\E_{p(\thetab|\data)}[h(\thetab)]\) as in
Equation~\ref{eq:expectation}. The argument \code{h} can be any python
\code{Callable}. The method \code{romc.eval_unnorm_posterior(theta,
  eps_cutoff = False)} computes the unnormalized posterior
approximation as in Equation~\ref{eq:approx_posterior}. The method \\
\code{romc.eval_posterior(theta, eps_cutoff = False)} evaluates the
normalized posterior estimating the partition function
\(Z = \int p_{d,\epsilon}(\thetab|\data)d\thetab\) using Riemann's
integral approximation. The approximation is computationally feasible
only in a low-dimensional parametric space.

\subsubsection*{Example - Inference part}

In the following code snippet, we use the inference utilities to (a)
get weighted samples from the approximate posterior, (b) compute an
expectation and (c) evaluate the approximate posterior. We also use
some of \pkg{ELFI}'s built-in tools to get a summary of the obtained
samples. For \code{romc.compute_expectation()}, we demonstrate its use
to compute the samples mean and the samples variance. Finally, we
evaluate \code{romc.eval_posterior()} at multiple points to plot the
approximate posterior of
Figure~\ref{fig:running_example_romc_inference}. We observe that the
approximation is quite close to the ground-truth.

\begin{Code}
# Inference part
seed = 21
n2 = 50
romc.sample(n2 = n2, seed = seed)

# visualize region, adding the samples now
romc.visualize_region(i = 1)

# Visualize marginal (built-in ELFI tool)
weights = romc.result.weights
romc.result.plot_marginals(weights = weights, bins = 100, range = (-3,3))

# Summarize the samples (built-in ELFI tool)
romc.result.summary()
# Method: ROMC
# Number of samples: 19300
# Parameter                Mean               2.5%              97.5%
# theta:                 -0.012             -1.985              1.987

# compute expectation
exp_val = romc.compute_expectation(h = lambda x: np.squeeze(x))
print("Expected value   : %.3f" % exp_val)
# Expected value: -0.012

exp_var = romc.compute_expectation(h = lambda x: np.squeeze(x)**2)
print("Expected variance: %.3f" % exp_var)
# Expected variance: 1.120

# eval unnorm posterior
print("%.3f" % romc.eval_unnorm_posterior(theta = np.array([[0]])))
# 58.800

# check eval posterior
print("%.3f" % romc.eval_posterior(theta = np.array([[0]])))
# 0.289
\end{Code}

\subsection{Evaluation part}
\label{subsec:evaluation}

The method \code{romc.compute_ess()} computes the Effective Sample
Size (ESS) as \(\frac{(\sum_i w_i)^2}{\sum_i w_i^2}\), which is a
useful quantity to measure how many samples actually contribute to an
expectation. For example, in an extreme case of a big population of
samples where only one has big weight, the ESS is much smaller than
the samples population.

\noindent
The method \code{romc.compute_divergence(gt_posterior, bounds, step,
  distance)} estimate the divergence between the ROMC approximation
and the ground truth posterior. Since the estimation is performed
using Riemann's approximation, the method can only work in low
dimensional spaces. The method can be used for evaluation in synthetic
examples where the ground truth is accessible. In a real-case
scenarios, where it is not expected to have access to the ground-truth
posterior, the user may set the approximate posterior obtained with
any other inference approach for comparing the two methods. The
argument \code{step} defines the step used in the Riemann's
approximation and the argument \code{distance} can be either
\code{"Jensen-Shannon"} or \code{"KL-divergence"}.

\begin{Code}
# Evaluation part
res = romc.compute_divergence(wrapper, distance = "Jensen-Shannon")
print("Jensen-Shannon divergence: %.3f" % res)
# Jensen-Shannon divergence: 0.035

nof_samples = len(romc.result.weights)
ess = romc.compute_ess()
print("Nof Samples: %d, ESS: %.3f" % (nof_samples, ess))
# Nof Samples: 19300, ESS: 16196.214
\end{Code}

\subsection{Extend the implementation with custom modules}
\label{subsec:extensibility}

ROMC is a generic LFI framework as it describes a sequence of steps
for approximating the posterior distribution without explicitly
enforcing a specific algorithm for each step. For completeness,
\cite{Ikonomov2019} propose a method for each step but, in general, a
user can experiment with alternative methods. Considering that, we
designed the implementation to support flexibility.

We have specified four critical parts where a user may intervene using
custom methods; (a) gradient-based optimization, (b) Bayesian
optimization, (c) proposal region construction and (d) surrogate model
fitting. Each of these parts corresponds to an internal function
inside the \code{romc.OptimisationProblem} class; (a)
\code{solve_gradients()}, (b) \code{solve_bo()}, (c)
\code{build_region()} and (d) \code{fit_local_surrogate()},
respectively. To replace any of these parts, the user must create a
custom class that inherits \code{OptimisatioProblem} and overwrite the
appropriate function(s).

To illustrate this in practice, suppose a user wants to fit Deep
Neural Networks instead of, the default, quadratic models as local
surrogates \(\tilde{d}_i\). Therefore, the user must create a new
class that inherits \code{OptimisationProblem} and overwrite the
\code{fit_local_surrogate(**kwargs)} function with one that fits
neural networks as local surrogates. We illustrate that in the
following snippet using the \code{neural_network.MLPRegressor} class
of the \pkg{scikit-learn} package. The reader can find the end-to-end
example
\href{https://colab.research.google.com/drive/1_jHVxPSH3XcNOORZJpLU0SPzs0PF8CQ5?usp=sharing}{here}
as an online Colab notebook.

\begin{Code}
class CustomOptim(OptimisationProblem):
    def __init__(self, **kwargs):
        super(CustomOptim, self).__init__(**kwargs)

    def fit_local_surrogate(self, **kwargs):
        nof_samples = 500
        objective = self.objective # the distance function

        # helper function
        def local_surrogate(theta, model_scikit):
            assert theta.ndim == 1
            theta = np.expand_dims(theta, 0)
            return float(model_scikit.predict(theta))

        # create local surrogate model as a function of theta
        def create_local_surrogate(model):
            return partial(local_surrogate, model_scikit = model)

        local_surrogates = []
        for i in range(len(self.regions)):
            # prepare dataset
            x = self.regions[i].sample(nof_samples)
            y = np.array([objective(ii) for ii in x])

            # train Neural Network
            mlp = MLPRegressor(hidden_layer_sizes = (10,10), solver = 'adam')
            model = Pipeline([('linear', mlp)])
            model = model.fit(x, y)
            local_surrogates.append(create_local_surrogate(model))

        self.local_surrogates = local_surrogates
        self.state["local_surrogates"] = True
\end{Code}

In a similar way, the user can replace any of the other three
functions. In each case, the custom function must update some
class-level variables that hold the state of the training phase. In
the following sections, we present which are these variables in each
function. Furthermore, when implementing custom functions, the user
may use two helping classes; (a) \code{RomcOpimisationResult}, that
stores the result of the optimization and (b) \code{NDimBoundingBox},
that stores the bounding box. We present their definitions in the
following snippets and we illustrate how to use them in the next
sections. Both classes can be imported from the module
\code{elfi.methods.inference.romc}.

\begin{Code}
class RomcOpimisationResult:
    def __init__(self, x_min, f_min, hess_appr):
        Parameters
        ----------
        x_min: np.ndarray (D,), minimum
        f_min: float, distance at x_min
        hess_appr: np.ndarray (D,D), Hessian approximation at x_min

class NDimBoundingBox:
    def __init__(self, rotation, center, limits):
        Parameters
        ----------
        rotation: np.ndarray (D,D), rotation matrix for the bounding box
        center: np.ndarray (D,) center of the bounding box
        limits: np.ndarray (D,2), the limits of the bounding box
\end{Code}

\subsubsection*{(a) Extending gradient-based optimization}

For replacing the default gradient-based optimization method, the user
must overwrite the function \code{solve_gradients()}. Using the
objective function \(d_i\) (\code{self.objective}), the custom method
must store the result of the optimization as a
\code{RomcOptimisationResult} instance. In the following snippet,
after the comment \code{\# state variables}, we present the
class-level variables that must be set by the method.

\begin{Code}
def solve_gradients(self, **kwargs):
    # useful variables
    # self.objective: Callable, the distance function

    # code custom solution here
    result = RomcOptimisationResult(x = .., y = .., jac = .., hess_inv = ..)
    success: bool = ... # whether optimization was successful

    # state variables
    self.state["attempted"] = True
    if success:
        self.result = result
        self.state["solved"] = True
        return True
    else:
        return False
\end{Code}

\subsubsection*{(b) Extending Bayesian Optimization}

For replacing the default Bayesian optimization function, the
procedure is similar to the gradient-based case. As presented in the
following snippet, the additional class-level variables that must be
set are; (a) \code{self.surrogate = custom_surrogate}, where
\code{custom_surrogate} is a \code{Callable} and (b)
\code{self.state["has_fit_surrogate"] = True} if the optimization is
successful.

\begin{Code}
def solve_bo(self, **kwargs):
    # useful variables
    # self.objective: Callable, the distance function

    # code custom solution here
    result = RomcOptimisationResult(x = .., y = .., jac = .., hess_inv = ..)
    custom_surrogate = ... # store a Callable here
    success: bool = ... # whether optimization was successful

    # state variables
    self.state["attempted"] = True
    if success:
        self.result = result
        self.surrogate = custom_surrogate
        self.state["solved"] = True
        self.state["has_fit_surrogate"] = True
        return True
    else:
        return False
\end{Code}

\subsubsection*{(c) Extending the proposal region construction}

For replacing the construction of the proposal region the user must
overwrite the \code{build_region} method. Using the objective function
\(d_i\) (\code{self.objective}) the method must estimate a list of
bounding boxes as a \code{List} with \code{NDimBoundingBox} instances
and set the state variables presented below. An end-to-end example for
using a custom region construction module can be found
\href{https://colab.research.google.com/drive/1RzB-V1QueP1y1nyzv_VOqR1nVz3DUH3v?usp=sharing}{here}.

\begin{Code}
def build_region(self, **kwargs):
    # useful variables
    # self.objective: Callable, the distance function

    # custom build_region method
    eps: float = ... # epsilon used in region estimation
    bounding_box: List[NDimBoundingBox] = ...
    success: bool = ... # whether region built successfully

    # state variables
    self.eps_region = eps
    if success:
        # construct region
        self.regions = bounding_box
        self.state["region"] = True
        return True
    else:
        return False
\end{Code}

\subsubsection*{(d) Extending the surrogate model fitting}

For replacing the surrogate model fitting the user must overwrite the
\code{fit_local_surrogate} method. Using the objective function
\(d_i\) (\code{self.objective}) and the estimated bounding boxes
(\code{self.regions}), the method must create a list of local
surrogates, one for each region, as a \code{List} with
\code{Callables} and set the state variables as presented in the
following snippet.

\begin{Code}
def fit_local_surrogate(self, **kwargs):
    # useful variables
    # self.objective: Callable, the distance function
    # self.regions: List[NDimBoundingBox], the bounding boxes

    # custom local surrogates
    local_surrogates: List[Callable] = ... # the surrogate models
    success: bool = ... # whether surrogates fit successfully

    # state variables
    if success:
        self.local_surrogates = custom_surrogates
        self.state["local_surrogates"] = True
        return True
    else:
        return False
\end{Code}

%% -- Use-case illustration ----------------------------------------------------
\section{Use-case illustration}

In this section, we test the implementation using the second-order
moving average (MA2) example, which is one of the standard models of
\pkg{ELFI}. We perform the inference using three different versions of
ROMC; (i) with a gradient-based optimizer, (ii) with Bayesian
Optimization and (iii) fitting a Neural Network as a surrogate
model. The later illustrates how to extend the implementation,
replacing part of ROMC with a user-defined component. Finally, we
measure the execution speed-up using the parallel version of ROMC.

\subsubsection*{Model Definition}

MA2 is a probabilistic model for time series analysis. The observation
at time \(t\) is given by,

\begin{gather} \label{eq:ma2}
y_t = w_t + \theta_1 w_{t-1} + \theta_2 w_{t-2}, \quad t=1, \ldots, T\\
\theta_1, \theta_2 \in \R, \quad  w_k \sim \mathcal{N}(0,1), k \in \mathbb{Z}
\end{gather}

\noindent
The random variable \(w_{k} \sim \mathcal{N}(0,1) \) is white noise
and the two parameters of interest, \(\theta_1, \theta_2\), model the
dependence from the previous observations. The parameter \(T\) is the
number of sequential observations which is set to \(T=100\). For
securing that the inference problem is identifiable, i.e., the
likelihood has only one mode, we use the prior proposed
by~\cite{Marin2012},

\begin{equation} \label{eq:ma2_prior}
p(\thetab) = p(\theta_1)p(\theta_2|\theta_1)
= \mathcal{U}(\theta_1;-2,2)\mathcal{U}(\theta_2;\theta_1-1, \theta_1+1)
\end{equation}

\noindent
The observation vector \(\yb_0 = (y_1, \ldots, y_{100})\) is generated
with \(\thetab^*=(0.6, 0.2)\). The dimensionality of the output
\(\yb\) is high, therefore we use summary statistics. Considering that
the output vector represents a time-series signal, we select the
autocovariances with \(\mathrm{lag}=1\) and \(\mathrm{lag}=2\), as
shown in Equations~\ref{eq:ma2_summary_1}
and~\ref{eq:ma2_summary_2}. The distance between the observation and
the simulator output is measured with the squared Euclidean distance,
as shown in Equation~\ref{eq:ma2_summary_4}.

\begin{gather}
  \label{eq:ma2_summary_1} s_1(\yb) = \frac{1}{T-1} \sum_{t=2}^T y_ty_{t-1}\\
  \label{eq:ma2_summary_2} s_2(\yb) = \frac{1}{T-2} \sum_{t=3}^T y_ty_{t-2} \\
  s(\yb) = (s_1(\yb), s_2(\yb))\\
  \label{eq:ma2_summary_4} d = ||s(\yb) - s(\yb_0)||_2^2
\end{gather}

\subsubsection*{Inference}

To demonstrate the full capabilities of our ROMC implementation, we
perform inference using three different methods: (i) a gradient-based
optimizer, (ii) Bayesian optimization, and (iii) fitting a neural
network (NN) as a surrogate model. The use of a NN as a surrogate
model serves as an example of the extensibility of our implementation,
as described in Chapter~\ref{subsec:extensibility}. For the NN, we
employ the \pkg{MLPRegressor} class from the \pkg{scikit-learn}
package. The NN (\(\tilde{d}_i\)) substitutes the actual distance
function (\(d_i\)) inside the proposal regions. Therefore, all
inference actions, namely, sampling, expectation computation, and
posterior evaluation are based on \(\tilde{d}_i\). We use a NN with
two hidden layers of 10 neurons each and train it using 500 examples
from each proposal region. To compare the results of ROMC inference
with a traditional ABC algorithm, we also include Rejection Sampling
in our analysis.

\begin{figure}[ht]
    \begin{center}
        \includegraphics[width=0.49\textwidth]{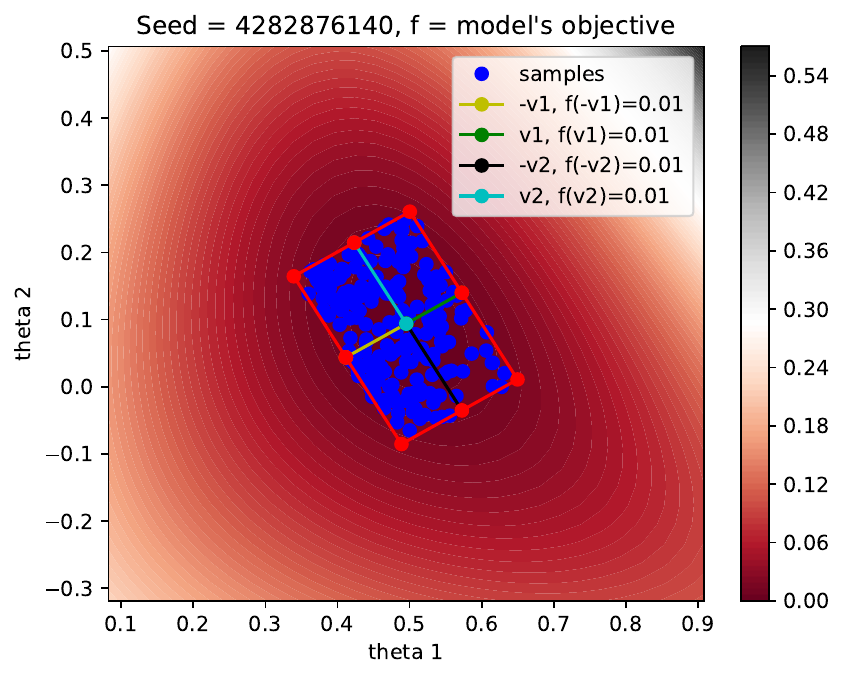}
        \includegraphics[width=0.49\textwidth]{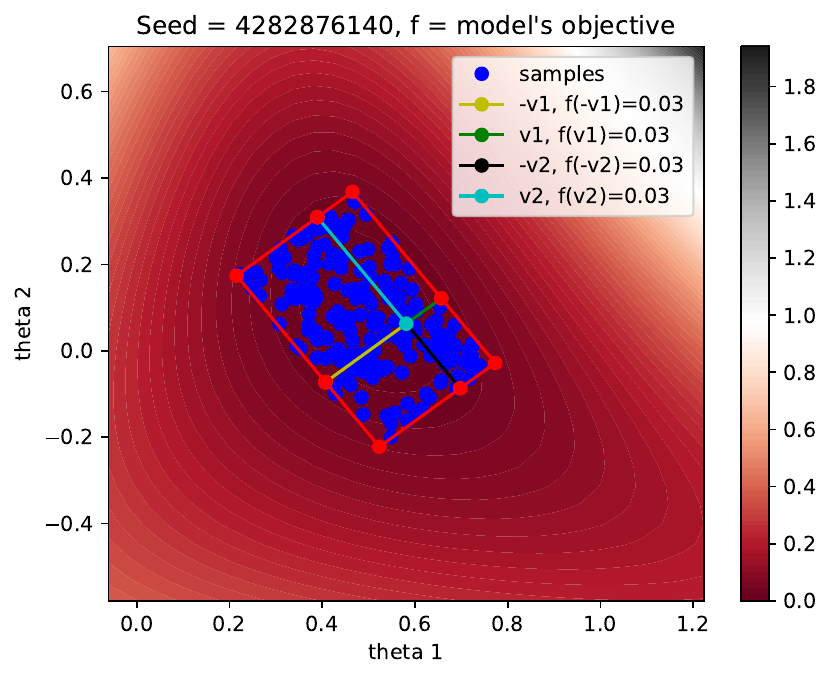}
    \end{center}
  \caption[The acceptance region of a specific deterministic simulator.]{The acceptance region in a specific optimization problem. In the left figure the region obtained with gradient-based optimizer and in the right one with Bayesian Optimization.}
  \label{fig:ma2_5}
\end{figure}

In Figure \ref{fig:ma2_5}, we illustrate the acceptance region of the
same deterministic simulator, in the gradient-based and the Bayesian
optimization case. The acceptance regions are quite similar even though the
different optimization schemes lead to different optimal points.

\begin{figure}[ht]
  \begin{center}
    \includegraphics[width=0.24\textwidth]{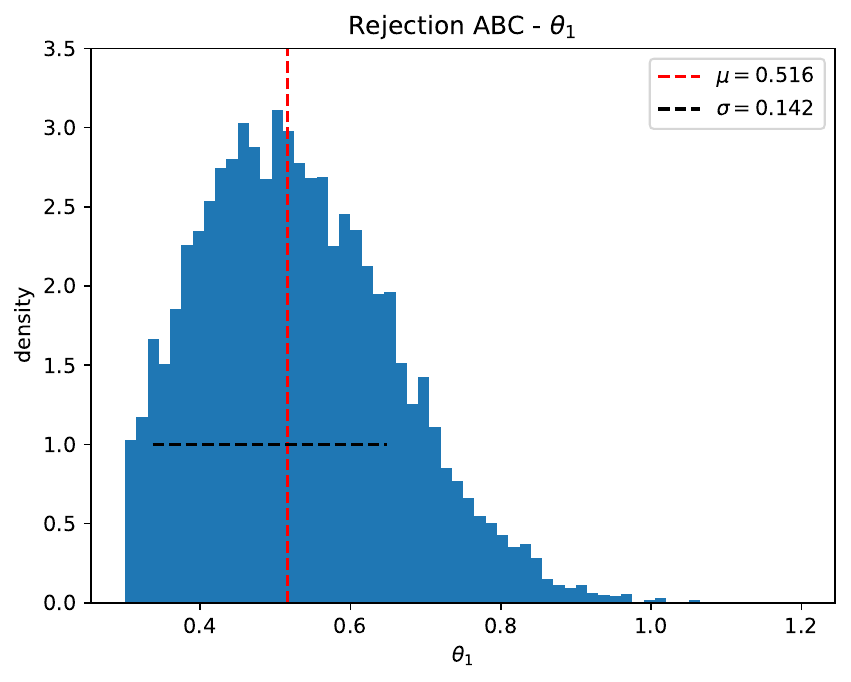}
    \includegraphics[width=0.24\textwidth]{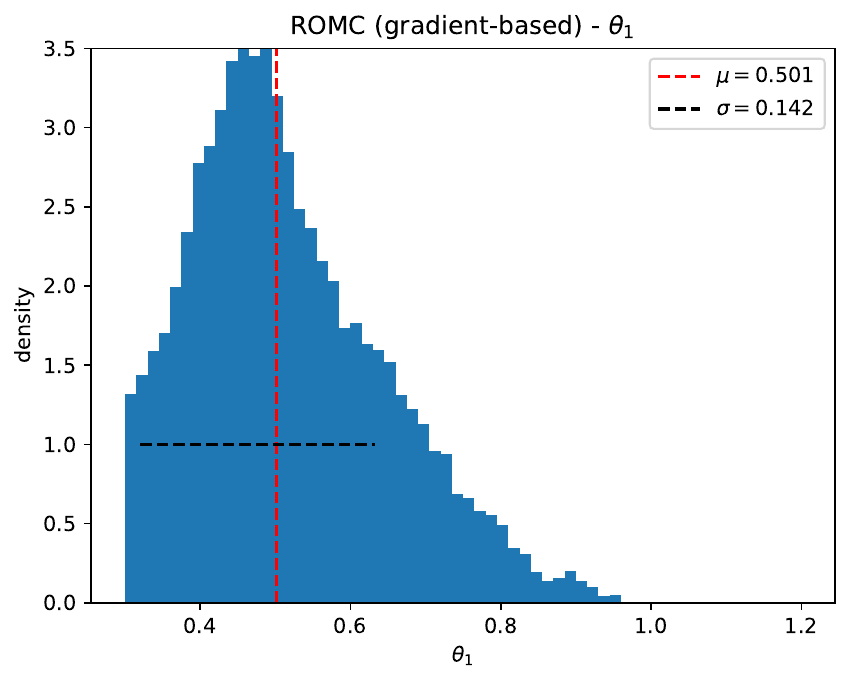}
    \includegraphics[width=0.24\textwidth]{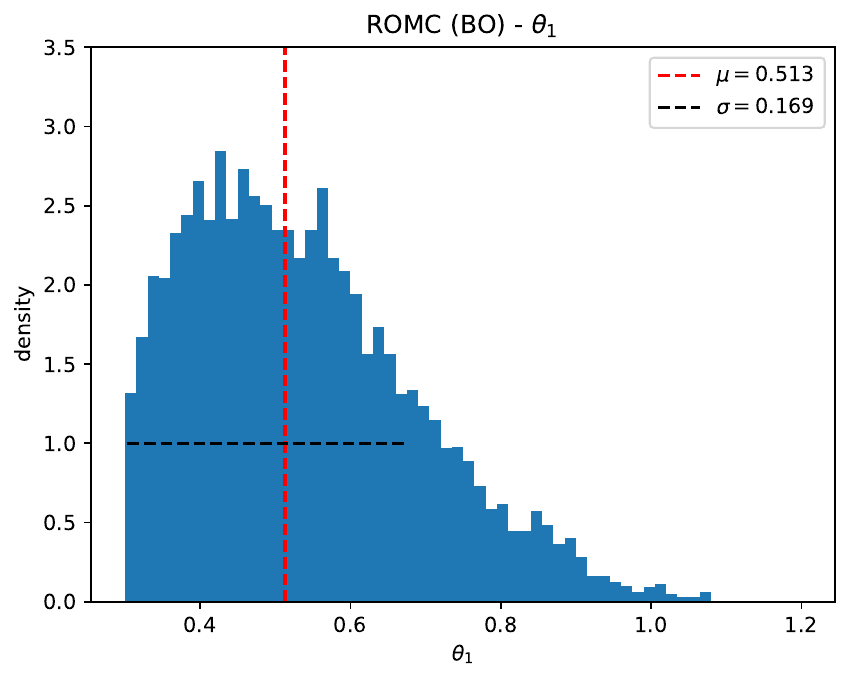}
    \includegraphics[width=0.24\textwidth]{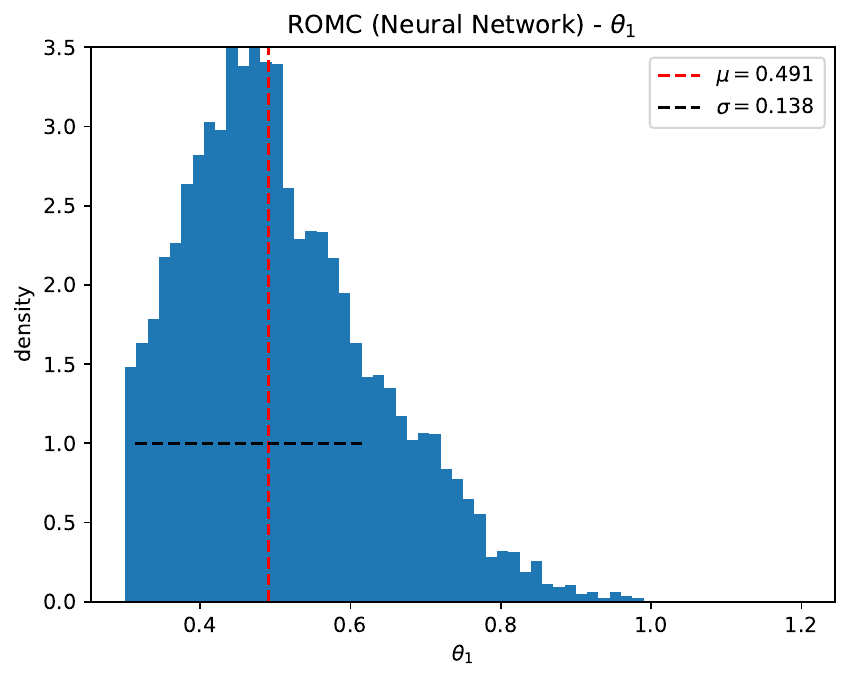}\\
    \includegraphics[width=0.24\textwidth]{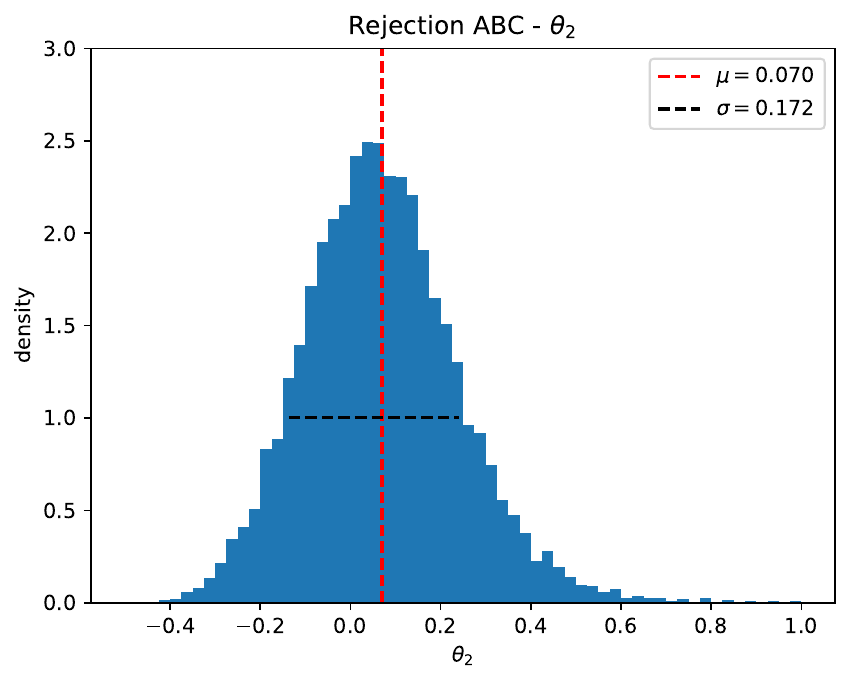}
    \includegraphics[width=0.24\textwidth]{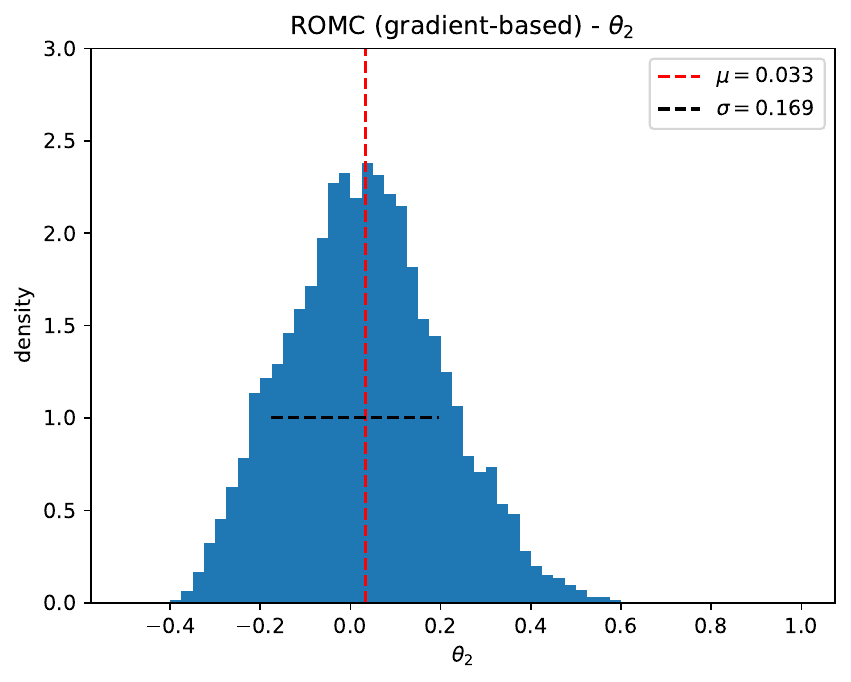}
    \includegraphics[width=0.24\textwidth]{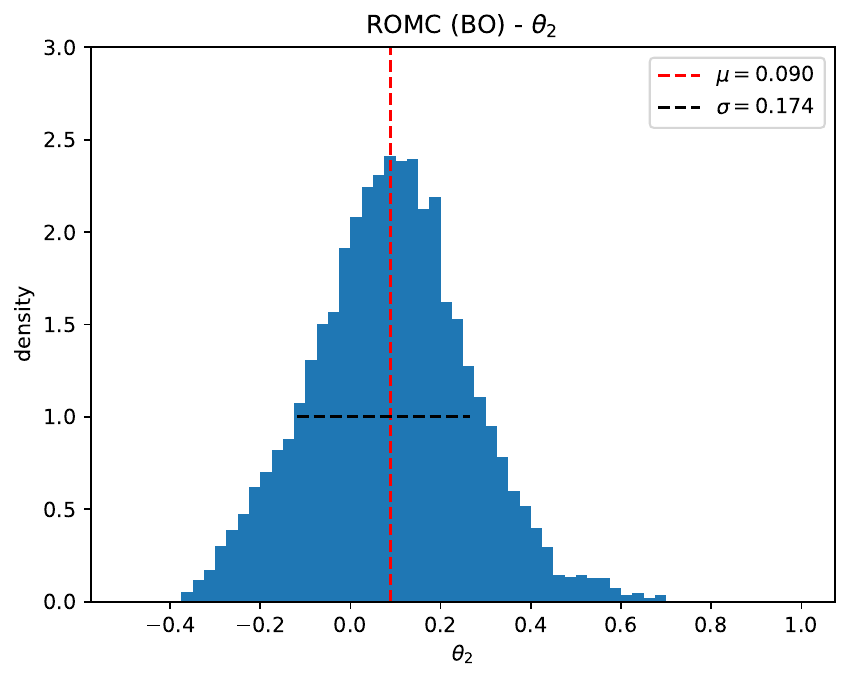}
    \includegraphics[width=0.24\textwidth]{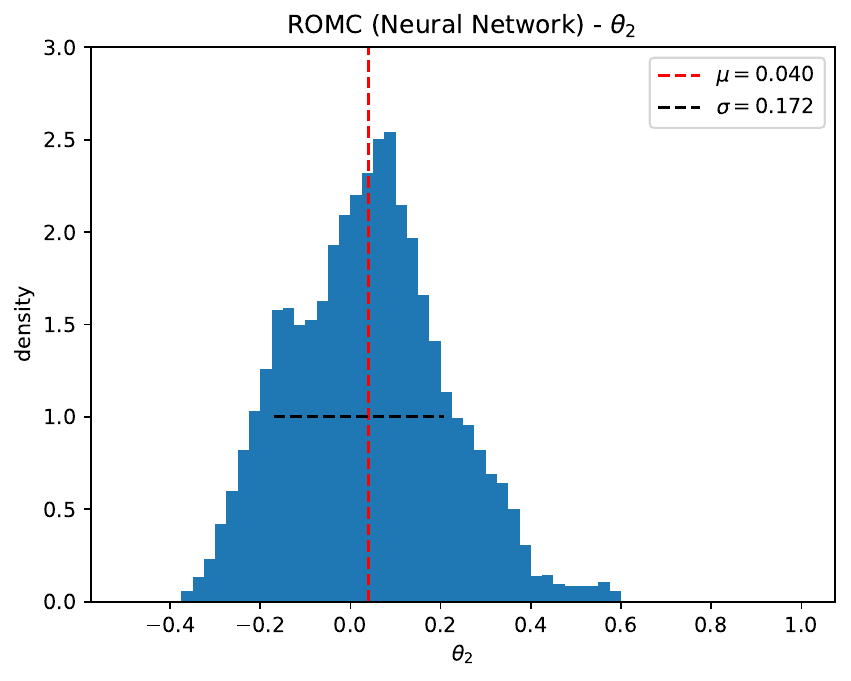}
    \end{center}
    \caption[MA2 example, evaluation of the marginal
    distributions.]{Histogram of the marginal posterior distributions
      using three different inference approaches; (a) in the first
      row, the samples are obtained using Rejection ABC sampling (b)
      in the second row, using ROMC with a gradient-based optimizer
      and (c) in the third row, using ROMC with Bayesian optimization
      approach. The vertical (red) line represents the samples mean
      \(\mu\) and the horizontal (black) the standard deviation
      \(\sigma\).}
  \label{fig:ma2_3}
\end{figure}

\begin{figure}[ht]
  \begin{center}
    \includegraphics[width=0.32\textwidth]{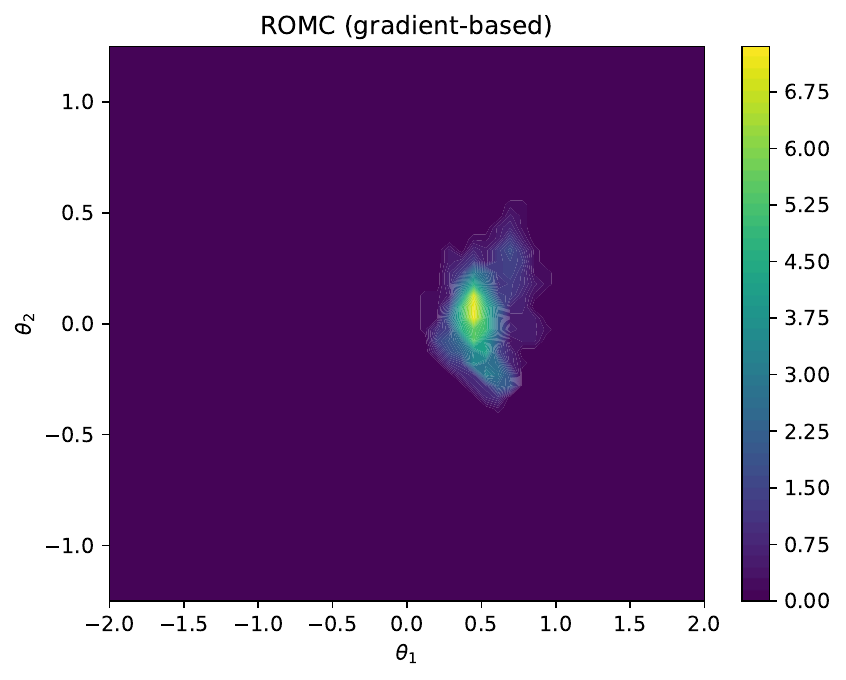}
    \includegraphics[width=0.32\textwidth]{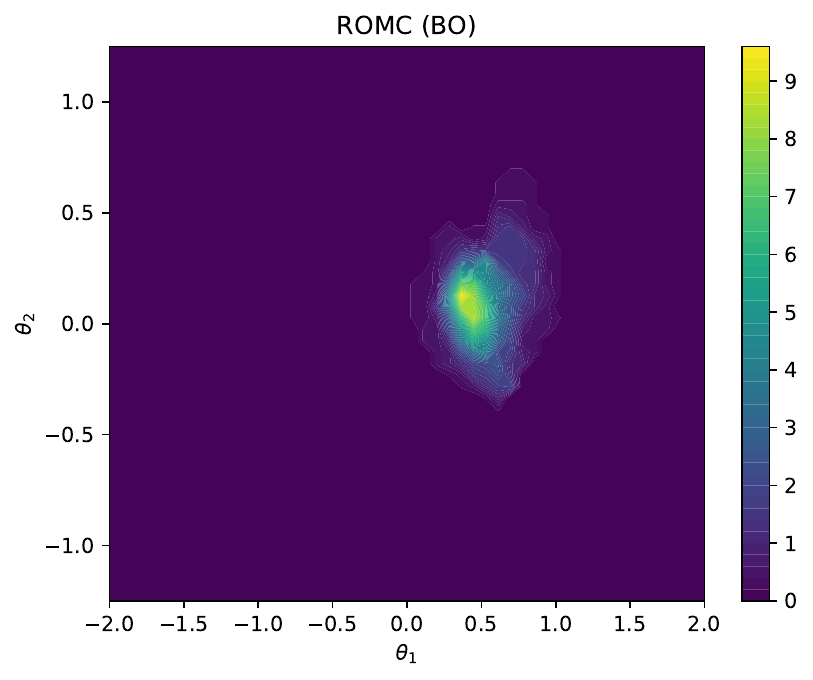}
    \includegraphics[width=0.32\textwidth]{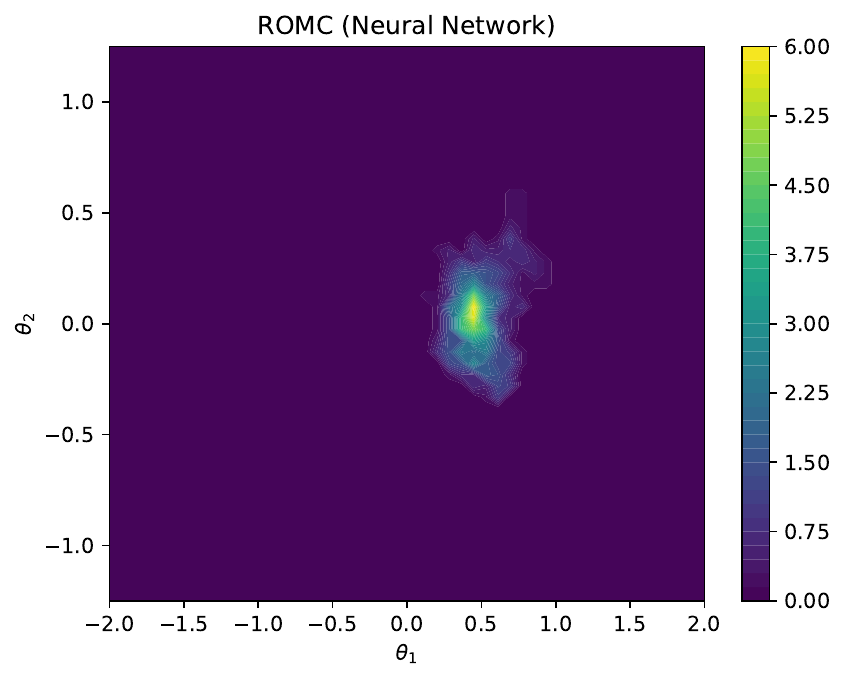}
    \end{center}
    \caption[MA2 example, posterior distribution.]{The unnormalized posterior distribution using the ROMC method with (a) a gradient-based optimization (b) Bayesian Optimization (c) gradient-based with a Neural Network as a surrogate model.}
  \label{fig:ma2_4}
\end{figure}

In Figure \ref{fig:ma2_3}, we demonstrate the histograms of the
marginal posteriors, for each approach; (a) Rejection ABC, (b) ROMC
with gradient-based optimization (c) ROMC with Bayesian optimization
and (d) ROMC with the NN extension. We observe a significant agreement
between the different approaches. At Table \ref{tab:ma2} we present
the empirical mean \(\mu\) and standard deviation \(\sigma\) for each
inference approach and finally, in Figure \ref{fig:ma2_4}, we
illustrate the unnormalized posterior for the three different
variations of the ROMC method. The results show that all ROMC
variations provide consistent results between them which are in
agreement with the Rejection ABC algorithm.

\begin{table}
\begin{center}
\begin{tabular}{ c|c|c|c|c }
\hline
& \(\mu_{\theta_1}\) & \(\sigma_{\theta_1}\) & \(\mu_{\theta_2}\) & \(\sigma_{\theta_2}\) \\
\hline \hline
Rejection ABC & 0.516 & 0.142 & 0.070 & 0.172 \\
\hline
ROMC (gradient-based) & 0.501 & 0.142 & 0.033 & 0.169 \\
\hline
ROMC (Bayesian optimization) & 0.513 & 0.169 & 0.090 & 0.174 \\
\hline
ROMC (Neural Network) & 0.491 & 0.138 & 0.040 & 0.172 \\
\hline
\end{tabular}
\end{center}
\caption{Comparison of the samples obtained from the estimated
  posterior with (a) Rejection sampling and (b) the different versions
  of ROMC. We observe that the obtained samples share similar
  statistics along all methods. \label{tab:ma2}}
\end{table}

\subsubsection*{Parallelize the implementation}

\begin{figure}[ht]
  \begin{center}
    \includegraphics[width=0.49\textwidth]{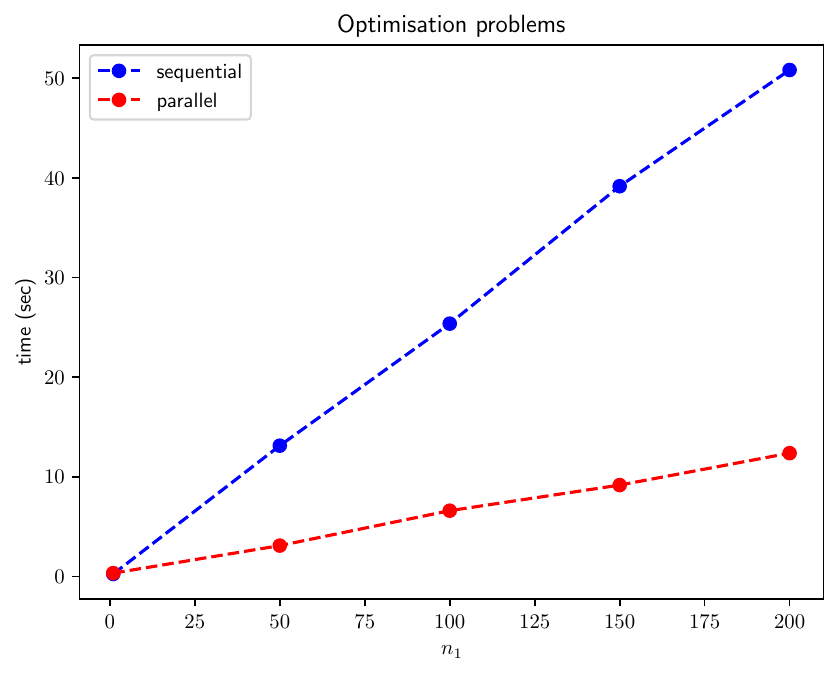}
    \includegraphics[width=0.49\textwidth]{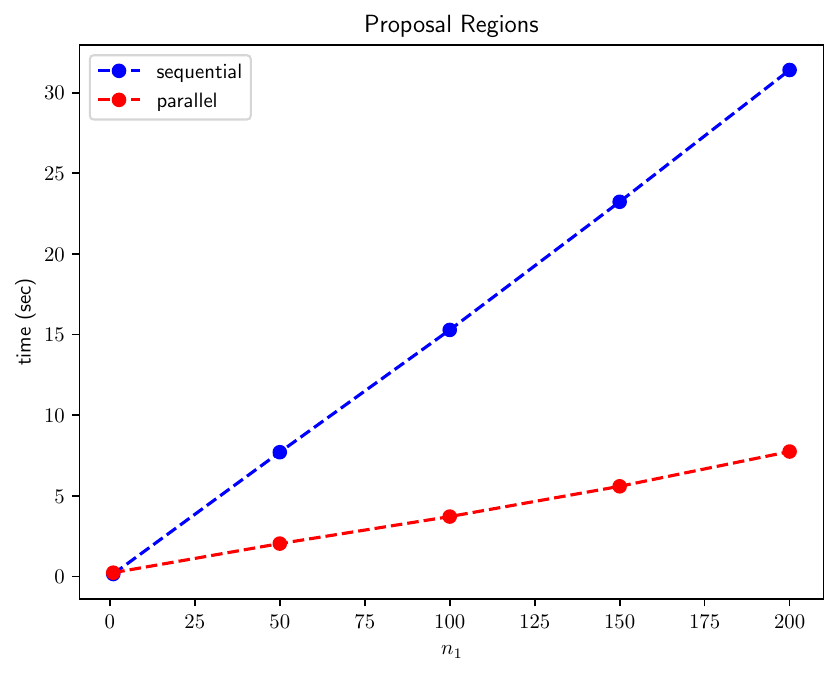}
    \end{center}
    \caption[Execution time exploiting Parallelize]{Comparison
      between parallel and sequential execution of ROMC. We observe
      that the parallel version runs almost 5 times faster.}
      \label{fig:exec_parallel}
\end{figure}

As stated above, ROMC is an approach that can be executed in a
fully-parallelized manner, exploiting all CPU cores. In our
implementation, we support a parallel version of the the training
part, namely, for solving the optimization problems and for estimating
the proposal regions. The parallel version of the algorithm is built
on top of the built-in \proglang{Python} package \pkg{multiprocess}
for using all the available CPU cores. In
Figure~\ref{fig:exec_parallel} we observe the execution times for
performing the inference on the MA2 model; the parallel version
performs both tasks almost five times faster than the sequential. The
result is reasonable given that the experiments have run in a single
machine with the Intel® Core™ i7-8750H Processor, which has six
separate cores.

%% -- Summary/conclusions/discussion -------------------------------------------
\section{Summary and discussion} \label{sec:summary}

In this paper, we presented the implementation of the LFI method ROMC
at the \proglang{Python} package \pkg{ELFI}. We highlighted two
different use-cases. First, we illustrated how a user may exploit the
provided API to solve an LFI problem. Second, we focus on the scenario
where a researcher wants to intervene and alter parts of the method to
experiment with new approaches. Since (Robust) Optimization Monte
Carlo is a novel approach for statistical inference and, to the best
of our knowledge, this is the first open-source implementation on a
generic package, we believe that the later is the biggest
contribution.

There are still open challenges for enabling ROMC to solve
high-dimensional LFI problems efficiently. The first is enabling the
execution of ROMC execution into a distributed environment, i.e., a
cluster of computers. ROMC can be characterized as an
\textit{embarrassingly parallel} workload; each optimization problem
is an entirely independent task. Therefore, supporting inference into
a cluster of computers can radically speed up the inference. The
second is the implementation of the method on a framework that supports
automatic differentiation. Automatic differentiation is necessary for
efficiently solving optimization problems, especially in
high-dimensional parametric models.

%% -- Optional special unnumbered sections -------------------------------------

\section*{Computational details}

The results in this paper were obtained using \proglang{Python~3.9},
\pkg{ELFI~0.8.6} at \proglang{Ubuntu 20.04 lts} operating system, at a
single machine with \proglang{Intel® Core™ i7-8750H} processor.

\section*{Acknowledgments}

HP was funded by European Research Council grant 742158 (SCARABEE,
Scalable inference algorithms for Bayesian evolutionary epidemiology).

\clearpage
\bibliography{refs}

%% -----------------------------------------------------------------------------

\end{document}